\documentclass[letterpaper, 10 pt, conference]{ieeeconf}  
\usepackage[T1]{fontenc}
\usepackage{ae,aecompl}

\usepackage{amssymb}
\usepackage{times}
\usepackage{caption}
\usepackage[labelformat=simple]{subcaption}
\usepackage{graphicx}
\usepackage{graphics}
\usepackage{blkarray}
\usepackage{booktabs}
\usepackage{xcolor} 
\usepackage{bbold}
\usepackage{caption}
\usepackage{subcaption}
\usepackage{multirow}
\usepackage{comment}
\usepackage{epstopdf}
\usepackage{tikz}
\usepackage{todonotes}
\graphicspath{{./figures/}}
\usepackage{csquotes}
\usepackage{siunitx}
\usepackage{cite}
\usepackage{epigraph}
\usepackage{empheq}
\usepackage{comment}
\usepackage{cancel}
\usepackage{algorithm} 
\usepackage{algorithmic}  
\usepackage{mathtools}
\usepackage{balance}
\usepackage{dirtytalk}

\usepackage[algo2e,linesnumbered,ruled,vlined]{algorithm2e}

\usepackage{hyperref}
\DeclareMathOperator*{\argmin}{\arg\!\min}

\newcommand{\bt}{\mathcal{T}}
\newcommand{\act}{\mathcal{A}}

\usepackage[T1]{fontenc}

\SetKwRepeat{Do}{do}{while}

\makeatother

\DeclareCaptionLabelSeparator{periodspace}{.\quad}
\IEEEoverridecommandlockouts                              
\newtheorem{remark}{Remark}

\newtheorem{example}{Example}

\newtheorem{definition}{Definition}
\newtheorem{experiment}{Experiment}

\makeatletter
\newcommand*{\rom}[1]{\expandafter\@slowromancap\romannumeral #1@}
\makeatother

\def\eqalignno#1{\let\\=\cr\displ@y \tabskip\@centering
  \halign to\displaywidth{\hfil$\@lign\displaystyle{##}$\tabskip\z@skip
    &$\@lign\displaystyle{{}##}$\hfil\tabskip\@centering
    &\llap{$\@lign##$}\tabskip\z@skip\crcr
    #1\crcr}}
\def\leqalignno#1{\let\\=\cr\displ@y \tabskip\@centering
  \halign to\displaywidth{\hfil$\@lign\displaystyle{##}$\tabskip\z@skip
    &$\@lign\displaystyle{{}##}$\hfil\tabskip\@centering
    &\kern-\displaywidth\rlap{$\@lign##$}\tabskip\displaywidth\crcr
    #1\crcr}}
\begin{document}

\thispagestyle{empty}
\twocolumn
\title{\LARGE \bf
Analysis and Exploitation of Synchronized Parallel Executions in Behavior Trees }

\author{Michele Colledanchise and Lorenzo Natale  
\thanks{The authors are with the iCub Facility, Istituto Italiano di Tecnologia. Genoa, Italy.
e-mail: {\tt{ michele.colledanchise@iit.it}} \newline This work was carried out in the context of the CARVE project, which has received funding from the European Union's Horizon 2020 research and innovation programme under grant agreement No 732410, in the form of financial support to third parties of the RobMoSys project.} }

 
\maketitle
\thispagestyle{empty}
\pagestyle{empty}

\begin{abstract}

Behavior Trees (BTs) are becoming a popular tool to model the behaviors of autonomous agents in the computer game and the robotics industry. One of the key advantages of BTs lies in their composability, where complex behaviors can be built by composing simpler ones. The parallel composition is the one with the highest potential since the complexity of composing pre-existing behaviors in parallel is much lower than the one needed using classical control architectures as finite state machines. However, the parallel composition is rarely used due to the underlying concurrency problems that are similar to the ones faced in concurrent programming.

In this paper, we define two synchronization techniques to tackle the concurrency problems in BTs compositions and we show how to exploit them to improve behavior predictability. Also, we introduce measures to assess execution performance, and we show how the design choices can affect them.

To illustrate the proposed framework, we provide a set of experiments using the R1 robot and we gather statistically-significant data.

\end{abstract}

%
%

\section{Introduction}
\label{sec:introduction}

Modeling robot's behaviors using Behavior Trees (BTs) is becoming an appreciated practice. Applications span from manipulation~\cite{rovida2017extended, zhang2019ikbt}
to non-expert programming~\cite{coronado2018development,paxton2017costar,shepherd2018engineering}. Other works include task planning \cite{neufeld2018hybrid}, learning~\cite{sprague2018adding, banerjee2018autonomous,hannaford2019hidden}, and  UAV systems~\cite{sprague2018improving, ogren}.
Using BTs, the designer creates robots behaviors by composing together actions and condition in a hierarchical fashion. There are different ways to create such compositions, each with its own semantic. A very powerful composition is the \emph{parallel} one, where the designer can easily encode the concurrent execution of several sub-behaviors. Unfortunately, the parallel composition is the least used composition due to its underlying concurrency problems~\cite{BTBook,colledanchise2018improving,rovida2018motion}. 

In this paper, we show how recent advances in BT compositions allow extending the use of BTs to those applications that either requires synchronized concurrent actions or predictable execution times. 

 The choice of using BTs to model robot behaviors is often driven by the fact that BTs are modular, flexible, and reusable~\cite{BTBook}. Moreover, they have also been shown
to generalize successful robot control architectures such
as the Subsumption Architecture~\cite{brooks1986robust} and the Teleo-Reactive
Paradigm~\cite{BTBook}.
Using BTs, the designer can compose simple behaviors using the so-called \emph{control flow nodes}. The most common control flow nodes, which will be described in this paper, are Sequence, Fallback, and Parallel.
The parallel execution of independent behaviors can arise several concurrency problems in any modeling language and BTs are no exception. However, the parallel composition of BTs is less sensitive to dimensionality problems than a classical finite state machine~\cite{BTBook}.

Real robots often require to execute behaviors concurrently, for efficiency or constraints satisfaction, as we will show in this paper. The presence of concurrent behaviors requires to face the same issues affecting concurrent programming. The solutions adopted in concurrent programming made a tremendous impact on software development, increasing their applicability and performance to the scale that it is now a common practice in modern software. 

Another issue faced in computer programming is predictability. Predictability is required in applications with safety-critical constraints. Similarly, we desire robots with predictable behaviors, especially at the developing stage, where actions may run with a different speed in the real world and in a simulation environment. Increasing predictability reduces the difference between simulated and real-world robots execution.

Concurrent behaviors are also well studied in the Human-Robot Interaction (HRI) community, where they show evidence of more \say{believable} robots' behaviors in the presence of coordinated movements~\cite{fischer2013impact} (e.g. the robot moves the arm to point at an object while moving the head to look at it.), coordinated robots and human movements~\cite{lee2011vision}, and coordinated gestures and diaogues~\cite{kopp2006towards}.

In our recent work~\cite{colledanchise2018improving}, we partially addressed the aforementioned issues by defining Concurrent BTs, where nodes expose information regarding progress and resource used and allow actions to have progress that depends on each other. In this paper, we extend our previous work by discriminating between absolute and relative synchronization techniques and show how we can exploit such synchronizations to improve behavior predictability. In addition, we introduce measures to assess execution performance and show how design choices affect them.  

The remainder of this paper is structured as follows: In Section~\ref{sec:related} we overview the related work. In Section~\ref{sec:background} we present the background. Then, in Section~\ref{sec:parallel}, we formulate two different synchronization techniques to coordinate actions and increase behavior predictability. In Section~\ref{sec:performance} we show the proposed performance measure and analyze their dependency on design choices. In Section~\ref{EE}, we provide an experimental evaluation with different use cases. We conclude the paper in Section~\ref{sec:conclusions}.

\newpage
\section{Related Work}
\label{sec:related}

The parallel composition has found relatively little use, compared to the other compositions, due to the intrinsic concurrency issues similar to the ones of computer programming such as race conditions and deadlocks. Current applications that make use of the parallel composition assume either that the BTs executed in parallel lie on orthogonal state spaces \cite{champandard2007enabling, rovida2017extended} or the that BTs executed in parallel have a predefined priority assigned~\cite{weber2010reactive} where, in conflict situations, it is executed only the BT with the highest priority. Other applications impose a mutual exclusion of actions in BTs that are executed in parallel if they have potential conflicts (e.g. sending commands to the same actuator)~\cite{BTBook} or 
they assume that BTs that are executed in parallel are not in conflict by design.

The parallel composition found large use in the BT-based task planner  \emph{A Behavior Language} (ABL)~\cite{mateas2002behavior} and in its further developments. ABL was originally designed for the game \emph{Fa\c{c}ade}, and it has received attention for
its ability to handle planning and acting at different deliberation layers, in particular, in Real-Time Strategy games~\cite{weber2010reactive}.
ABL executes sub-BTs in parallel and resolves conflicts between multiple concurrent actions by defining a fixed priority order. This solution threatens the reusability and modularity of BTs and introduces the risk of starvation (i.e. the execution of an action with low priority may be perpetually denied to execute an action with higher priority).

The parallel composition found use also in multi-robot applications~\cite{colledanchise2016advantages}
where a single robot
BT is extended to a so-called \emph{multi-robot} BT, resulting in improved fault tolerance
and other performances. 
The parallel node involves multiple robots, each assigned to a specific task using a task-assignment algorithm. The task-assignment algorithm ensures the absence of conflicts.

None of the work above addressed properly the synchronization issues that arise when using a parallel BT node.

A recent work~\cite{rovida2018motion} proposed BTs for executing actions
in parallel, even when they lie on the same state space (e.g.
they use the same robot arm). The coordination mechanism is conducted by activating and deactivating motion primitives based on their conditions. Such framework avoids that more actions access a critical resource concurrently. In our work, we are interested in synchronizing the progress of actions that can be executed concurrently.

In our recent work~\cite{colledanchise2018improving}, we address the aforementioned issue by defining BT nodes that expose information regarding progress and resource uses. We also defined a relative synchronized parallel BT node execution and we provided theoretical validation of the proposed nodes. In this paper, we extend our previous work by defining absolute relative synchronization and we define and  study their performance.

To conclude, there is currently no work in exploiting and analyzing the performance of the synchronized parallel node. This makes our paper fundamentally different than the ones presented above and the BT literature.

\section{Background}
\label{sec:background}

In this section, we briefly present the classical formulation of BTs and we introduce the concepts for synchronization barriers and predictability. A more detailed description of BTs is available in~\cite{BTBook} while a more detailed description of concurrent programming is available in~\cite{taubenfeld2006synchronization}.
\subsection{Behavior Trees}
\label{sec:background.BT}

A BT is a graphical modeling language used as a representation for actions orchestration. A BT is as a directed rooted tree where the internal nodes represent behavior compositions and leaf nodes represent actuation or sensing operations. 

The children of a BT node are placed below it, as in Figure~\ref{EE:fig:timeline:BT}, and they are executed in the order from left to right. The execution of a BT begins from the root node. It sends \emph{ticks}, which are activation signals, with a given frequency to its children. A node in the tree is executed if and only if it receives ticks. When the node no longer receives ticks, its execution is aborted.  The child returns to the parent a status, which can be either \emph{Success}, \emph{Running}, or \emph{Failure} according to the node's logic. Below we present the most common BT nodes and their logic.


\paragraph*{Fallback}
When a Fallback node receives ticks, it sends ticks to its own children in order from the left. It returns a status of Success or Running whenever it finds a child that returns Success or Running respectively. It returns Failure whenever all the children return Failure. When a child returns Running or Success, the Fallback node does not send ticks to the next child (if any). 
The Fallback node is represented by a square with the label \say{$?$}, as in Figure~\ref{EE:fig:timeline:BT}.

\paragraph*{Sequence}
When a Sequence node receives ticks, it sends ticks to its own children in order from the left. It returns Failure or Running whenever it finds a child that returns Failure or Running respectively. It returns Success whenever all the children return Success. When a child returns Running or Failure, the Sequence node does not send ticks to the next child (if any). 
The Sequence node is graphically represented by a square with the label \say{$\rightarrow$}, as in Figure~\ref{EE:fig:timeline:BT}.

\paragraph*{Parallel}
The Parallel node sends ticks to all its children. It returns Success if all children return Success, it returns Failure if at least one child returns Failure, and it returns Running otherwise.
The parallel node is graphically represented by a square with the label \say{$\rightrightarrows$}.

\paragraph*{Action}
Whenever an Action node receives ticks, it performs some operations. It returns Success whenever the operations are completed and Failure if the operations cannot be completed. It returns Running otherwise. When a running Action no longer receives ticks, its execution is aborted.
An Action node is graphically represented by a rectangle, as in Figure~\ref{EE:fig:timeline:BT}.

\paragraph*{Condition}
Whenever a Condition node receives ticks, it checks if a proposition is satisfied or not. It returns Success or Failure accordingly. A Condition is graphically represented by an ellipse as in Figure~\ref{EE:fig:timeline:BT}.

The state-space formulation of BTs~\cite{BTBook} was defined to study them from a mathematical standpoint. In that formulation, the tick is represented by a recursive function call that includes both the return status, the system dynamics, and the system state. This formulation allows us to define concepts of \emph{progress} and \emph{safeguarding BTs}, used in this paper. 


\begin{definition}[Behavior Tree \cite{BTBook}]
\label{bg.def:BT}
A BT is a three-tuple 
\begin{equation}
 \bt_i=\{f_i,r_i, \Delta t\}, 
\end{equation}
where $i\in \mathbb{N}$ is the index of the tree, $f_i: \mathbb{R}^n \rightarrow  \mathbb{R}^n$ is the right hand side of a difference equation, $\Delta t$ is a time step and 
$r_i$ is the return status that can be equal to either \emph{Running}, \emph{Success}, or \emph{Failure}. Finally,  let $x_k=x(t_k)$ be the system state at time $t_k$, then the execution of a BT $\bt_i$ is described by the following equations:
\begin{eqnarray}
 x_{k+1}&=&f_i( x_{k}),  \label{bts:eq:executionOfBT}\\
 t_{k+1}&=&t_{k}+\Delta t.
\end{eqnarray}
 \end{definition}

\begin{definition}[Progress Function \cite{colledanchise2018improving}]
\label{bg.def.progress}
The function $p: R^n \to [0,1]$ is the progress function. It indicates the progress of the BT's execution at each state.
\end{definition}

\begin{definition}[Safeguarding \cite{BTBook}]
\label{bg.def.safeguarding}
 A BT is Safeguarding, with respect to the step length $d$, the obstacle region $O \subset \mathbb{R}^n$, and the initialization region $I \subset R$,
 if it is safe, and finite time successful with region of attraction $R' \supset I$~\cite{BTBook} and a success region $S$ such that $I$ surrounds $S$ in the following sense:
\begin{equation}
  \{x\in \mathbb{R}^n: \inf_{s\in S_1} || x-s  || \leq d \} \subset I.
\end{equation}
\end{definition}

\subsection{Concurrent Programming}

\label{sec:background.cp}

Concurrent programming deals with the execution of several concurrent processes that need to be synchronized to achieve a task or simply to avoid being in conflict with one another. The main uses of synchronization are \emph{producer-consumer relationship}, where a consumer process has to wait until a producer provides the necessary data, and \emph{exclusive use of resources}, where multiple processes have to use or access a critical resource and a correct synchronization strategy ensures that only one process at a time can access the resource~\cite{taubenfeld2006synchronization}.
The use of \emph{barriers} is one popular way to implement correct synchronization strategies~\cite{taubenfeld2006synchronization}.
The use of barriers allows concurrent processes to wait for each other at a specific point of execution.

\subsection{Predictability}

Predictability represents the ability to ensure the execution of an application without concern that outside factors will affect it in unpredictable ways. In other words, the application will behave as intended in terms of functionality, performance, and response time.

Predictability is highly appreciated in real-time systems. A real-time system must behave in a way that can be predicted in time to estimate the likelihood that a task can be completed before a given deadline.


\section{Parallel Synchronization of BT}
\label{sec:parallel}

In this section, we present the first contribution of this paper. We extend our previous work on parallel synchronization of BTs~\cite{colledanchise2018improving} by introducing \emph{absolute} and \emph{relative} synchronized parallel nodes and we show how they can be used to synchronize actions with both durative (with a given duration and progress) and perpetual (without a given duration and progress) actions.  Moreover, we show how to  use these synchronization techniques to improve the predictability of the progress of a BT.

\subsection{Absolute Synchronized Parallel Node}
\label{PM:AS}

Absolute synchronization is achieved by setting, a-priori, a finite ordered set $\mathcal{B}$ of values for the progress. These values are used as \emph{barriers} at the task level (see Section~\ref{sec:background.cp}). Whenever a node in this parallel composition has the progress equal to or greater than a progress barrier in $\mathcal{B}$
it no longer receives ticks until all the other nodes of the parallel composition have the progress equal to or greater than the value of that the barrier. 

Algorithm~\ref{ps.alg.asynch} shows the pseudocode of the absolute synchronized parallel node. At each tick, the node first assesses the minimum progress of its children (Lines 2-3), then it finds, among the predefined barriers contained in $\mathcal{B}$, the \emph{current barrier} (i.e. the barrier of smaller value that the progress of at least one child has not reached) (Lines 4-7). Then it sends ticks only to those actions whose progress did not exceed the value of the current barrier (Lines 8-10). Finally, it computes its return status (Lines 11-15). 
The absolute synchronized parallel node is graphically represented by a square with the label \say{$\rightrightarrows^A$}.

\begin{algorithm2e}[t!]
\SetKwProg{Fn}{Function}{}{}

\Fn{Tick()}
{

  \For{$i \gets 1$ \KwSty{to} $N$}
  {
  \ArgSty{minProgress} $\gets$ \FuncSty{min}(\ArgSty{minProgress}, $p_i$)
  }

  \For{$b  \in \mathcal{B}$}
  {
    \If{$b >$\ArgSty{minProgress}}
    {
    \ArgSty{current-barrier} $\gets b$\\
        \bf{break}
    }        
  }
  \ForAll{$i \gets 1$ \KwSty{to} $N$}
  {

  \If{$p_i \leq$ \ArgSty{current-barrier}}{
    \ArgSty{childStatus}[i] $\gets$ \ArgSty{child($i$)}.\FuncSty{Tick()}\\
  }
    }
    \uIf{$\Sigma_{i: \ArgSty{childStatus}[i]=Success}1 = N $}
    {
      \Return{Success}
    }
    \ElseIf{$\Sigma_{i: \ArgSty{childStatus}[i]=Failure}1 > 0$}
    {
      \Return{Failure}
    
  }
  \Return{Running}
  
  }
    \caption{Pseudocode of an absolute synchronized parallel node with $N$ children.}
 \label{ps.alg.asynch}
\end{algorithm2e}

We now present a use case example for the absolute synchronized parallel node.

\begin{example}[Door Pulling]
A humanoid robot has to pull a door open. This task requires the synchronization of two actions, the arm movement to pull the door and the mobile base movement to make the robot move away from the door while this opens. To succeed in the task, the arm movement and the mobile base must be synchronized.
\end{example}

\subsection{Relative Synchronized Parallel Node}
\label{PM:RS}
Relative synchronization is achieved by setting a-priori a threshold value $\Delta \in [0, 1]$. Whenever a node in this parallel composition has a progress that exceeds the minimum progress, among all the other nodes of the parallel composition,  by $\Delta$, it no longer receives ticks.

Algorithm~\ref{ps.alg.rsynch} shows the pseudocode of the relative synchronized parallel node. At each tick, the node first assesses the minimum progress of its children (Lines 2-4), then it sends ticks only to those actions whose progress did not exceeds the minimum progress by the predefined offset $\Delta$  (Lines 5-7). Finally, it computes its return status (Lines 8-12). 

The relative synchronized parallel node is graphically represented by a square with the label \say{$\rightrightarrows^R$}.
\begin{algorithm2e}[t!]
\SetKwProg{Fn}{Function}{}{}

\Fn{Tick()}
{
\ArgSty{minProgress} $\gets$ 1 \\ 
  \For{$i \gets 1$ \KwSty{to} $N$}
  {
  \ArgSty{minProgress} $\gets$ \FuncSty{min}(\ArgSty{minProgress}, $p_i$)
  }
  \ForAll{$i \gets 1$ \KwSty{to} $N$}
  {

  \If{$p_i \leq$ \ArgSty{minProgrees} + $\Delta$}{
    \ArgSty{childStatus}[i] $\gets$ \ArgSty{child($i$)}.\FuncSty{Tick()}\\
  }
    }
    \uIf{$\Sigma_{i: \ArgSty{childStatus}[i]=Success}1 = N $}
    {
      \Return{Success}
    }
    \ElseIf{$\Sigma_{i: \ArgSty{childStatus}[i]=Failure}1 > 0$}
    {
      \Return{Failure}
    
  }
  \Return{Running}
  
  }
    \caption{Pseudocode of a relative synchronized parallel node with $N$ children.}
 \label{ps.alg.rsynch}
\end{algorithm2e}

We now present a use case example for the relative synchronized parallel node.

{\begin{example}[Relative]
A service robot has to give directions to visitors of a museum. To make the robot's motions look natural, whenever the robot gives a direction, it points with its arm to the direction while moving the head to such direction. The arm and head may require different times to perform the motion. By imposing a relative progress synchronization we avoid the unnatural behavior where the robot looks first to a direction and then points at it, or the other way round.    
\end{example}



%

The relative synchronized parallel node can be used also to impose coordination between \emph{perpetual} action, (i.e. an action that even in the ideal case, do not have a fixed duration, hence a progress profile), as in the following example.

\begin{example}[Perpetual Actions]
A service robot has to carry around different tools in a workshop and it uses a cart to carry them. This behavior can be described as the relative parallel BT composition of two actions: one for holding the cart straight, and one for navigation. While the robot is pushing the cart forwards, the cart may drift sideways, just like any ordinary cart. Since the navigation and manipulation actions are executed concurrently in two independent actions the robot may move too fast and the cart may drift away before the robot can align it. By setting the progress of both actions to $1$ whenever the error of, respectively, arm or base reference position is within a boundary and $0$ otherwise, the base movements stops while the robot is aligning the cart. 
\end{example}
We will present the implementation of the example above in Section~\ref{EE}.

\begin{remark}
Designing a single action that operates both arm and base represents an easier synchronized solution. However, creating the single action for composite behaviors reduces the reusability of single behaviors and the whole BT.
\end{remark}

\subsection{Improving Predictabity}
\label{PM:IP}
Progress synchronization can be used to impose a given progress profile constraint. The idea is to define an artificial action with the desired progress profile (over time) defined a priori and putting it as a child of an absolute synchronized parallel node with the actions whose progress is to be constrained. 
However, since we are allowed to only stop actions (i.e. BTs have no means to speed up actions), we can only define such artificial action as the ideal upper bounds of the other actions progresses. 

\begin{example}
An industrial robot has to perform several manipulation tasks. Depending on the tool used, the movements have to take different progress profile. 
\end{example}
We will present the implementation of the example above in Section~\ref{EE}.

\begin{remark}

This type of progress-profile creation may become very useful at the developing stage, since the actions may run with a different speed in the real world and in a simulation environment. Improving predictability reduces the difference between simulated and real-world robots execution.
\end{remark}

\section{Performance Analysis of Synchronized BTs}
\label{sec:performance}

In this section, we present the second contribution of this paper. We define measures for the concurrent execution of BTs used to establish execution performance. We show measures for both progress synchronization and predictability.  We also show how the design choices for relative and absolute parallel nodes affect the performance.


\subsection{Progress Synchronization Distance}

\begin{definition}
\label{PM:def:performance}
The progress distance over a time window $[k_1,k_2]$ for a parallel node with $N$ children is defined as:

\begin{equation}
\pi(k_1,k_2) \triangleq  \sum_{k = {k_1}} ^{{k_2}} \sum_{i = 1} ^N{\sum_{j = 1}^N{\frac{|p_i(x_k) - p_j(x_k)|}{2}}}
\end{equation}
where $p_i\in [0, 1]$ is the progress of the $i$-th child, as in Definition~\ref{bg.def.progress}. 
\end{definition}

Intuitively, a small progress distance results in high performance for both relative and absolute nodes synchronization.

%
%
%
%

\subsection{Predictability Distance}
\label{pm.subsec:timeline}

A useful method to measure predictability is to set the desired progress value and compute the average variation from the expected and the  true time instant in which the action has a progress that is closest to the desired one.\footnote{It could be the case that the progress is defined only at discrete points of execution.} This measure can be used to assess the deviation from the ideal execution.

\begin{definition}
Let $\bar p \in [0,1]$ and  $T^{\bar p}$ be the set of time instances $t_k$ such that $t_k = \argmin{(p(x(t_k)) - \bar p)}$, collected by running a BT a finite number of times. The time predictability distance relative to progress $\bar p$ is defined as:
\begin{equation}
P(\bar p) \triangleq mean(T^{\bar p}) - \bar t_k
\end{equation}
\end{definition}
where $\bar t_k$ is the time instance when $p(x(t_k))$ is expected to be equal to $\bar p$.

%
%
%
%



\subsection{Sensitivity Analysis}
\label{sa}
We are ready to show how the number of barriers in $\mathcal{B}$ (for absolute synchronization) and the threshold value $\Delta$ (for relative synchronization) affect the performance, computed using the measures defined in this section. For illustrative purposes, we define custom made actions with difference progress profiles. To collect statistically-significant data, we ran the BT of each example 1000 times and we use plot boxes to compactly show the minimum, the maximum, the median, and the interquartile range of the measures proposed.

\subsubsection{How the number of equidistant barriers affects the performance of absolute synchronization}
We now present an example that highlights how the number of progress barriers in $\mathcal{B}$  affects the performance of absolute synchronization. In the example, we consider equidistant progress barriers.
\begin{example}
\label{PM.ex.dummy}
    Let a BT $\bt$ be an absolute parallel synchronization with the actions $\act_1$, $\act_2$, and $\act_3$ as children. The actions are such that the progress profile of each $\act_i$ holds Equation~\eqref{SA:ex:progress} below:
\begin{equation}
    p_i(x_k)= 
\begin{cases}
    0 &\text{ if }k = 0\\
    p_i(x_{k-1}) + \alpha_i + \omega_i(x_k),              & \text{otherwise}
\end{cases}
\label{SA:ex:progress}
\end{equation}

with $\alpha_i$s for each action $\act_i$ as: $\alpha_1 = 1$, $\alpha_2 = 2$, and $\alpha_3 = 5$; $\omega_i(x_k) \in [-\bar \omega, \bar \omega]$ a number, sampled from an uniform distribution, in the interval $[-\bar \omega, \bar \omega]$.

The model above describes an action whose progress evolves with a fixed value ($\alpha_i$) and with some noise ($\omega_i(x_k)$), modeling possible uncertainties in the execution that affect the progress. 

Figure~\ref{EE:fig:example5} shows the results of running  1000 times the BT in Example~\ref{PM.ex.dummy} in different settings. We observe higher performance with a large number of barriers and smaller  $\bar \omega$. This highlights that a higher number of barriers prevents the progress of the actions to differ from each other (see Algorithm~\ref{ps.alg.asynch} Line~5 and 9-10). Moreover, a larger $\bar \omega$ results in an higher increase in the progress between one tick and the next one, resulting in worse performance.   
\end{example}

\begin{figure}[h!]
\begin{subfigure}[b]{0.49\columnwidth}
\includegraphics[width=\columnwidth]{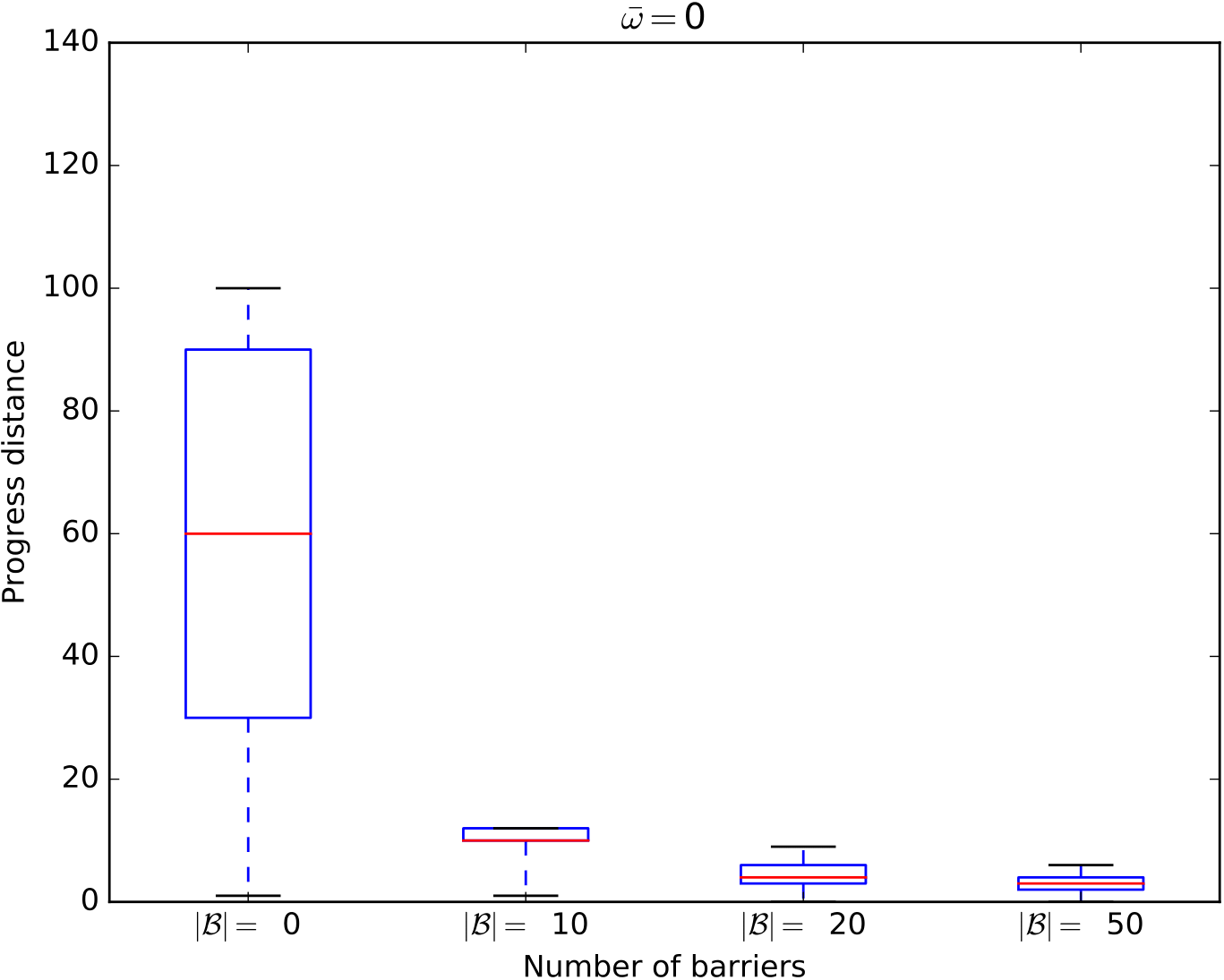}
\caption{Progress distances with $\bar \omega = 0$.}
\end{subfigure}
\begin{subfigure}[b]{0.49\columnwidth}
\includegraphics[width=\columnwidth]{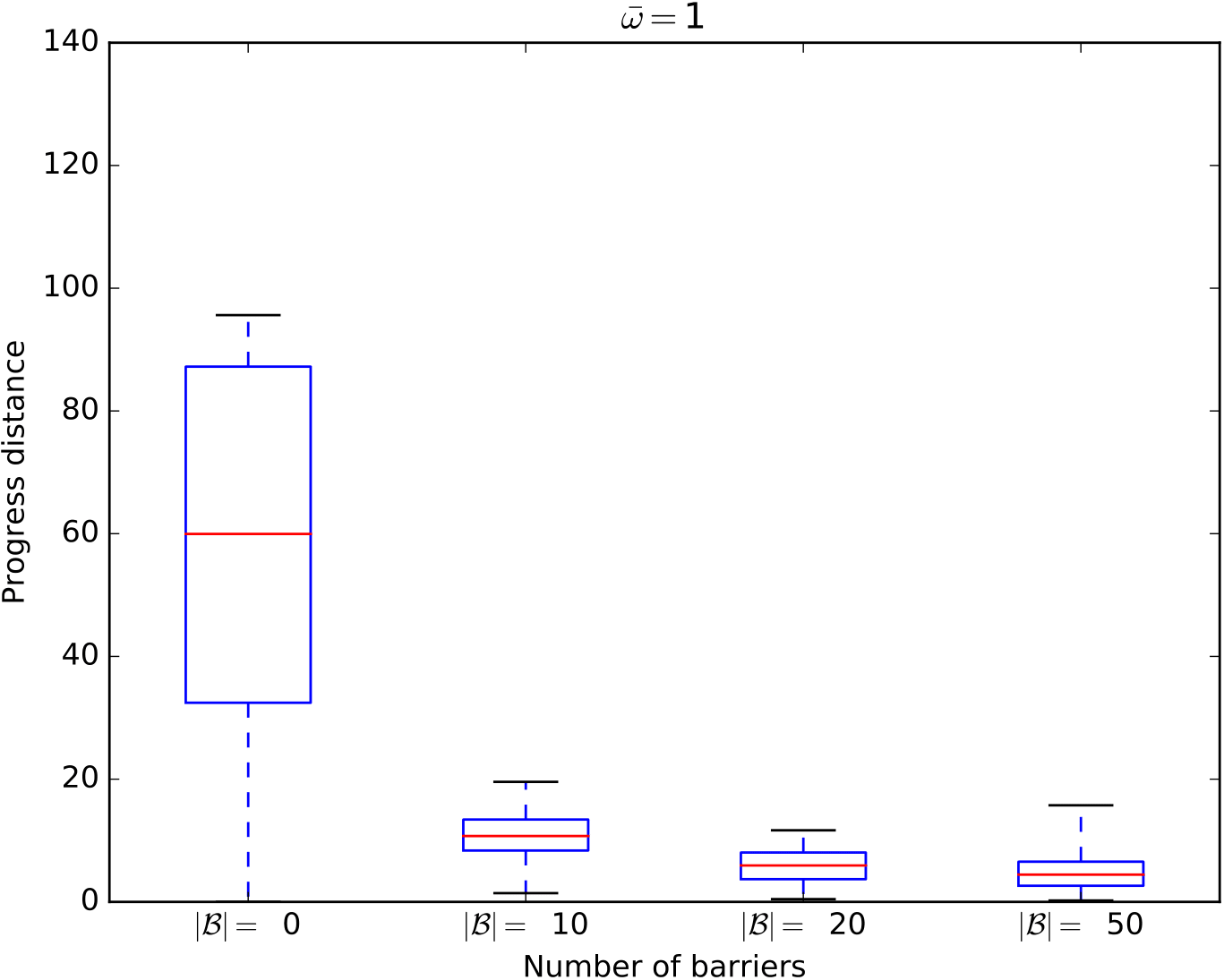}
\caption{Progress distances with $\bar \omega = 1$.}
\end{subfigure}
\vspace*{1em}

\begin{subfigure}[b]{0.49\columnwidth}
\includegraphics[width=\columnwidth]{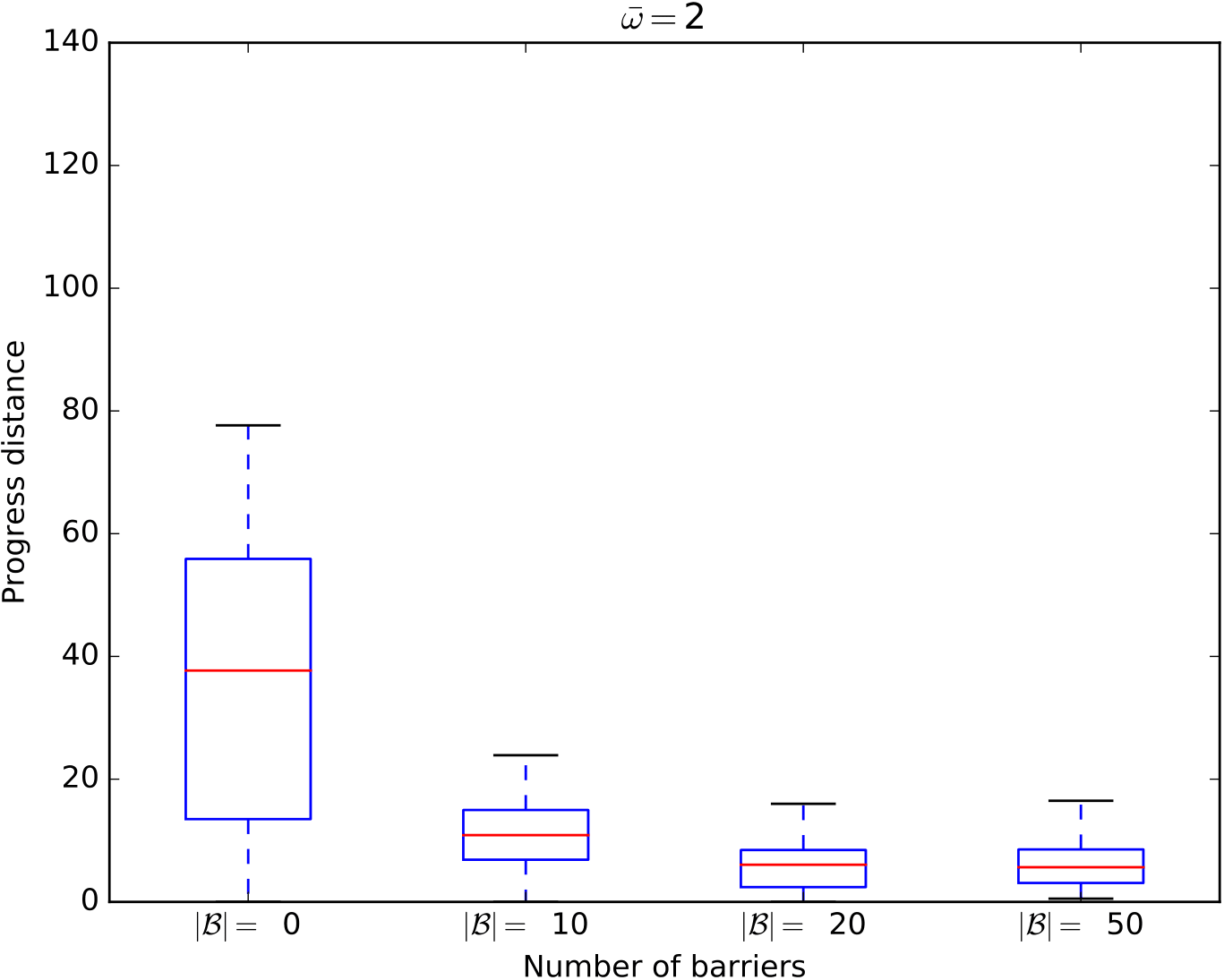}
\caption{Progress distances with $\bar \omega = 2$.}
\end{subfigure}
\begin{subfigure}[b]{0.49\columnwidth}
\includegraphics[width=\columnwidth]{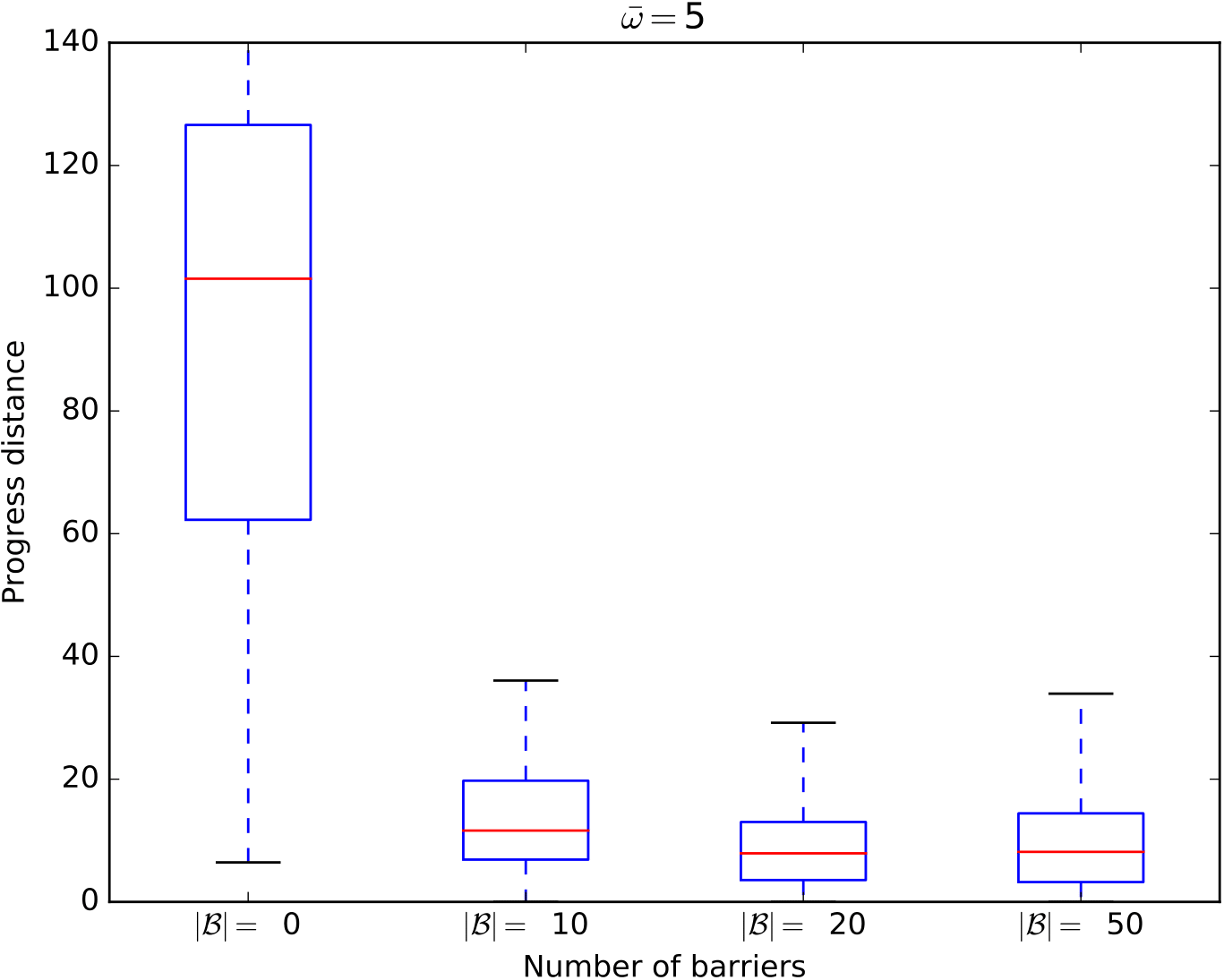}
\caption{Progress distances with $\bar \omega = 5$.}
\end{subfigure}
\caption{Plotbox for progress distance of Example~\ref{PM.ex.dummy} with different number of barriers $|\mathcal{B}|$. $|\mathcal{B}|= 0$ refers to the unsynchronized parallel execution.}
\label{EE:fig:example5}
\end{figure}

%

\subsubsection{How the threshold value affects the performance of relative synchronization}
We now present an example that highlights how the value of $\Delta$ affects the performance of relative synchronization.

\begin{example}
\label{PM.ex.dummy.delta}
Let a BT $\bt$ be a relative parallel synchronization with the same actions of Example~\ref{PM.ex.dummy} as children. Figure~\ref{PM:fig:cart} shows the results of running  1000 times the BT in Example~\ref{PM.ex.dummy.delta} in different settings.
 We observe that the performance increases with a smaller $\Delta$ and decreases with a larger $\bar \omega$. This highlights that a smaller $\Delta$ prevents the progress of the actions to differ from each other (see Algorithm~\ref{ps.alg.rsynch} Lines~3-7). Moreover, a larger $\bar \omega$ result in a higher possible difference in the progress of two actions between one tick and the next one, resulting in worse performance.
\begin{figure}[h!]
\begin{subfigure}[t]{0.49\columnwidth}
\includegraphics[width=\columnwidth]{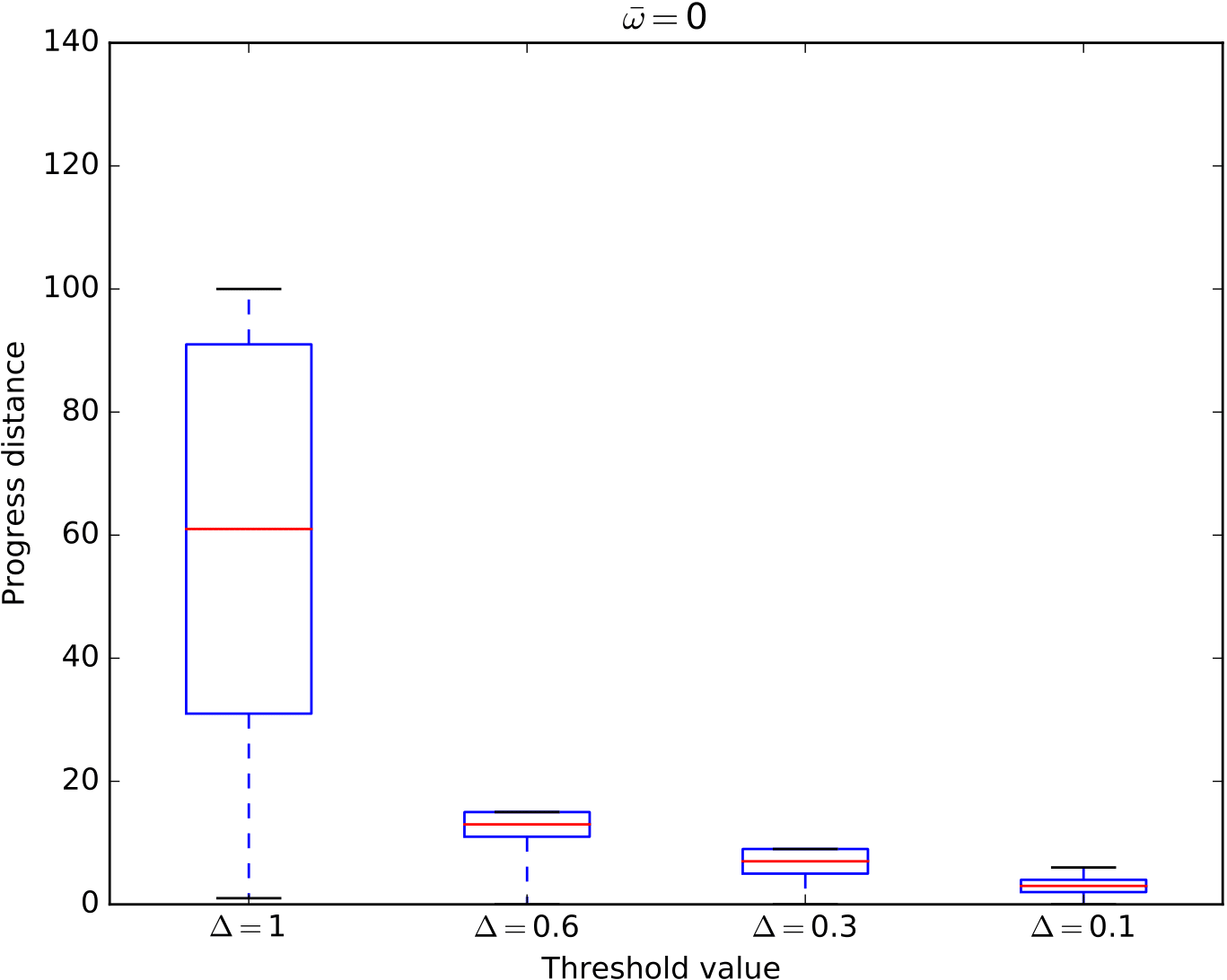}
\caption{Progress distances with $\bar \omega = 0$.}
\end{subfigure}
\begin{subfigure}[t]{0.49\columnwidth}
\includegraphics[width=\columnwidth]{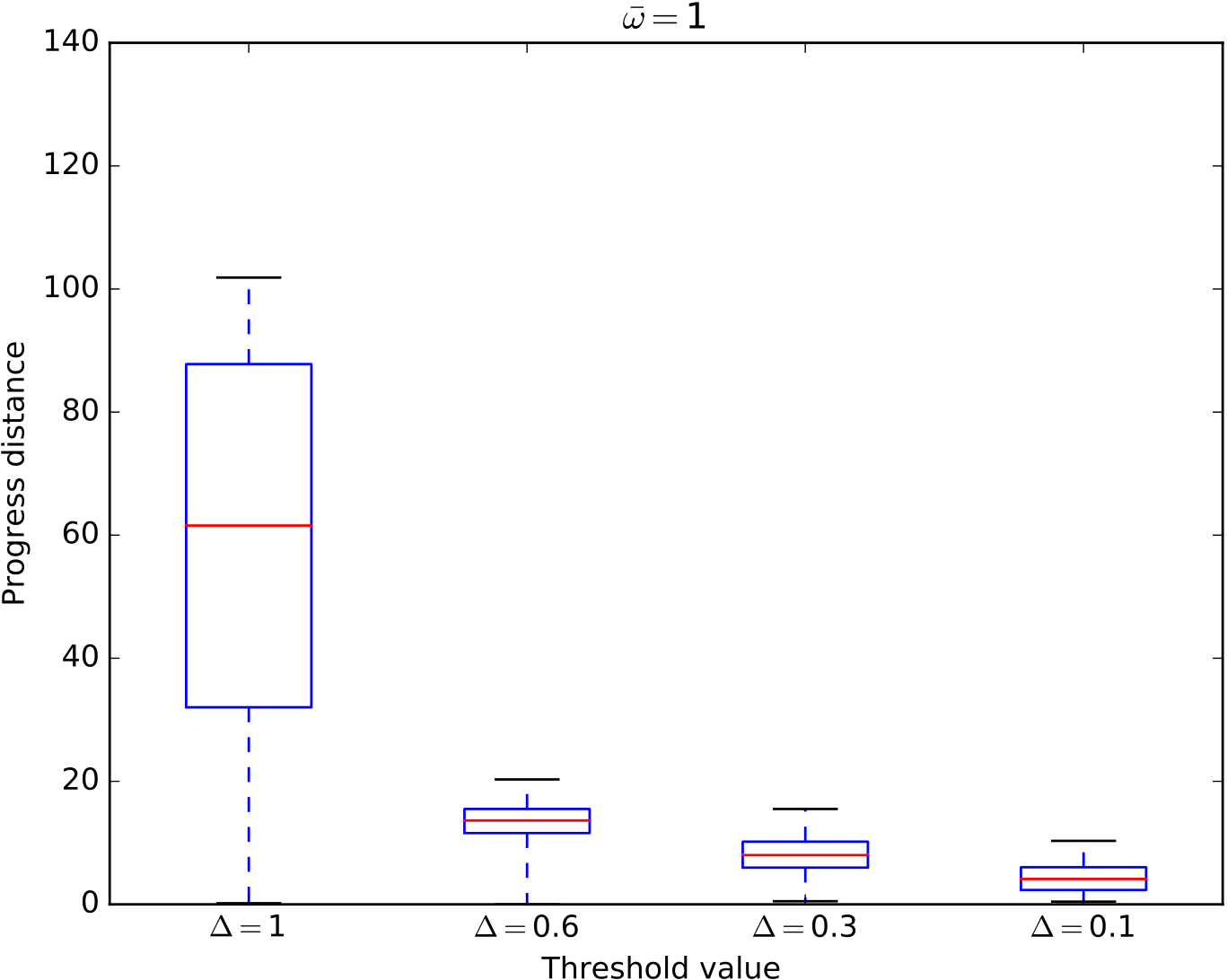}
\caption{Progress distances with $\bar \omega = 1$.}
\end{subfigure}

\vspace*{0.5em}
\begin{subfigure}[t]{0.49\columnwidth}
\includegraphics[width=\columnwidth]{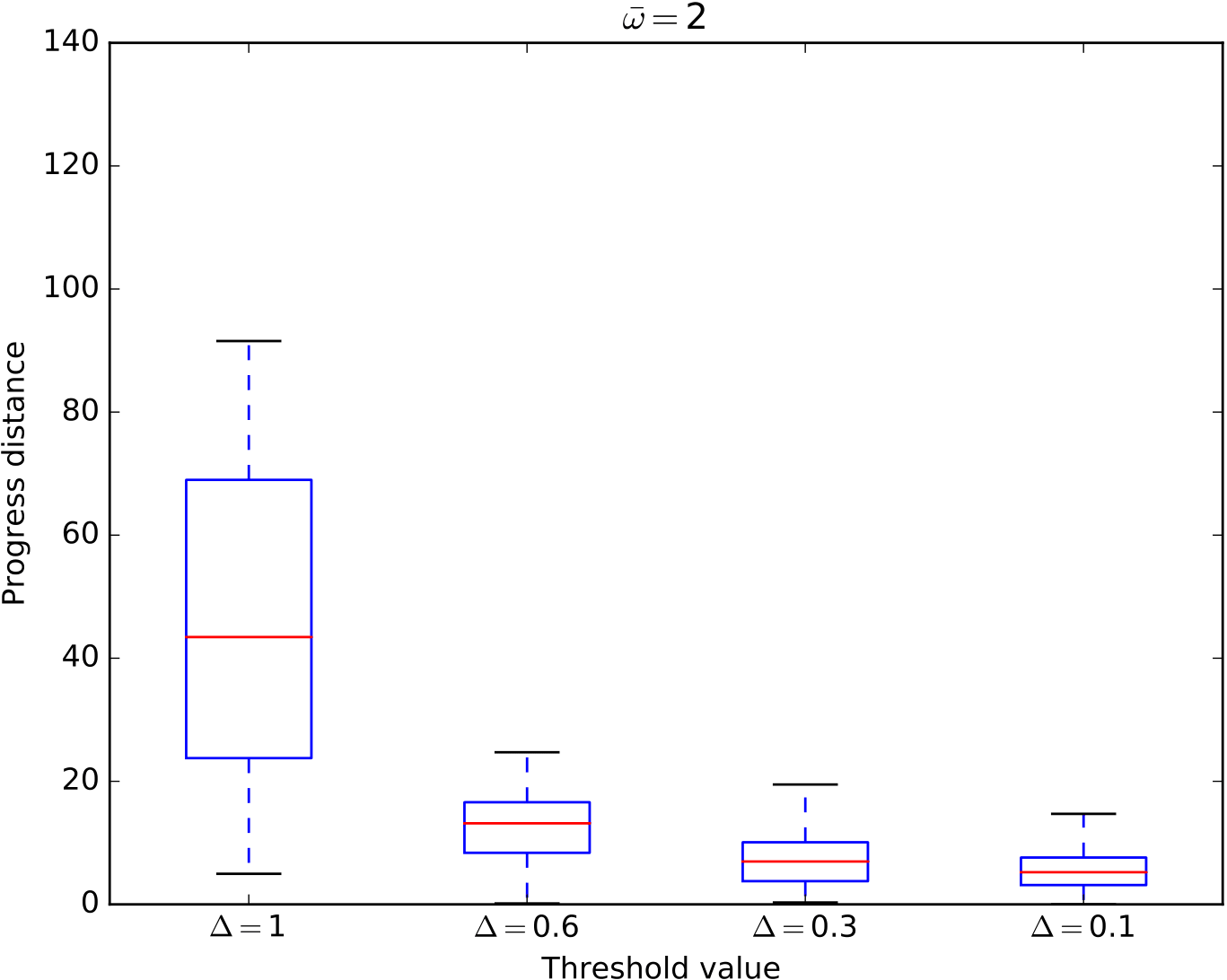}
\caption{Progress distances with $\bar \omega = 2$.}
\end{subfigure}
\begin{subfigure}[t]{0.49\columnwidth}
\includegraphics[width=\columnwidth]{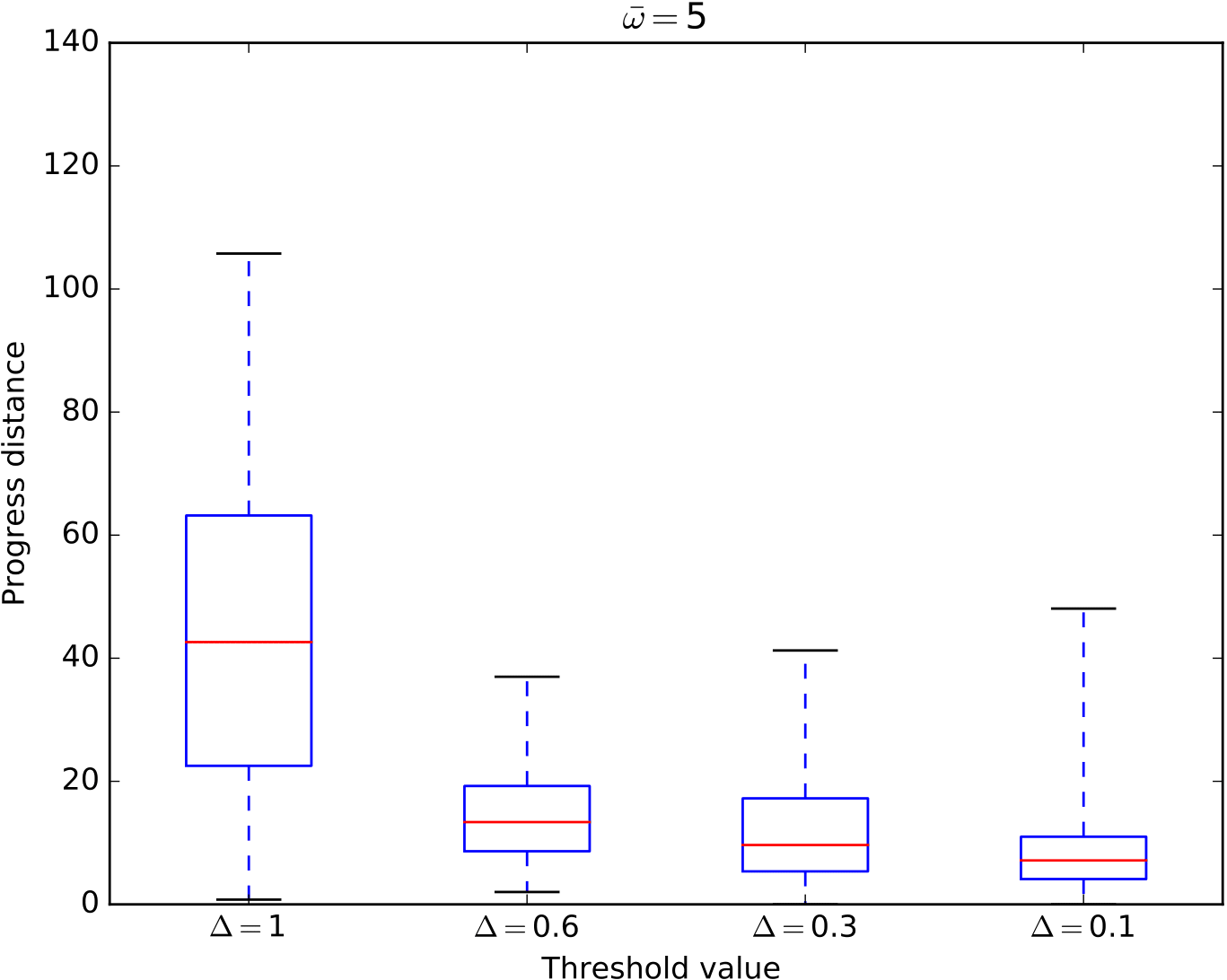}
\caption{Progress distances with $\bar \omega = 5$.}
\end{subfigure}
\caption{Plotbox for progress distance of Example~\ref{PM.ex.dummy.delta} with different $\Delta$. $\Delta = 1$ refers to the unsynchronized parallel execution.}
\label{PM:fig:cart}
\end{figure}

\end{example}
\clearpage

\begin{remark}
The synchronization may deteriorate other desired qualities. For example, since actions are waiting for one another, the overall execution may be slower than the slowest action. Moreover, a small value for $\Delta$  or a larger number of barriers can result in highly intermittent behaviors.
\end{remark}

\subsubsection{How the number of barriers affects the predictability}
We now present two examples that highlights how the number of barriers for an absolute synchronized parallel node affects the predictability of an execution.
\begin{example}
\label{PM.ex.dummy.timeline}
Let a BT $\bt$ be an absolute parallel synchronization that the actions $\act_1$ and $\act_2$ as children. $\act_1$ is an artificial action that has the desired progress profile over time (see Section~\ref{PM:IP}), whose progress holds Equation~\eqref{SA:ex:jitter:model} below:
\begin{equation}
    p_1(x_k)= 
\begin{cases}
    0 &\text{ if }k = 0\\
    p_1(x_{k-1}) + 0.1,              & \text{otherwise}
\end{cases}
\label{SA:ex:jitter:model}
\end{equation}
Hence, the desired progress profile is such that it starts as $0$ and it increases by $0.1$ at each time step. $\bt_2$ is a the action whose progress is to be imposed. Without constraints, the progress of $\bt_2$ holds Equation~\eqref{SA:ex:jitter:action} below:
\begin{equation}
    p_2(x_k)=  
\begin{cases}
    0 &\text{ if }k = 0\\
    p_2(x_{k-1}) + 2 + \omega_i(x_k),              & \text{otherwise}
\end{cases} 
\label{SA:ex:jitter:action}
\end{equation}
\end{example}

Figure~\ref{PM:fig:timeline} reports the results of Example~\ref{PM.ex.dummy.timeline}. We observe worse performance with larger $\bar \omega$ and $\Delta$.

\begin{figure}[h!]
%

\begin{subfigure}[t]{0.49\columnwidth}
\includegraphics[width=\columnwidth]{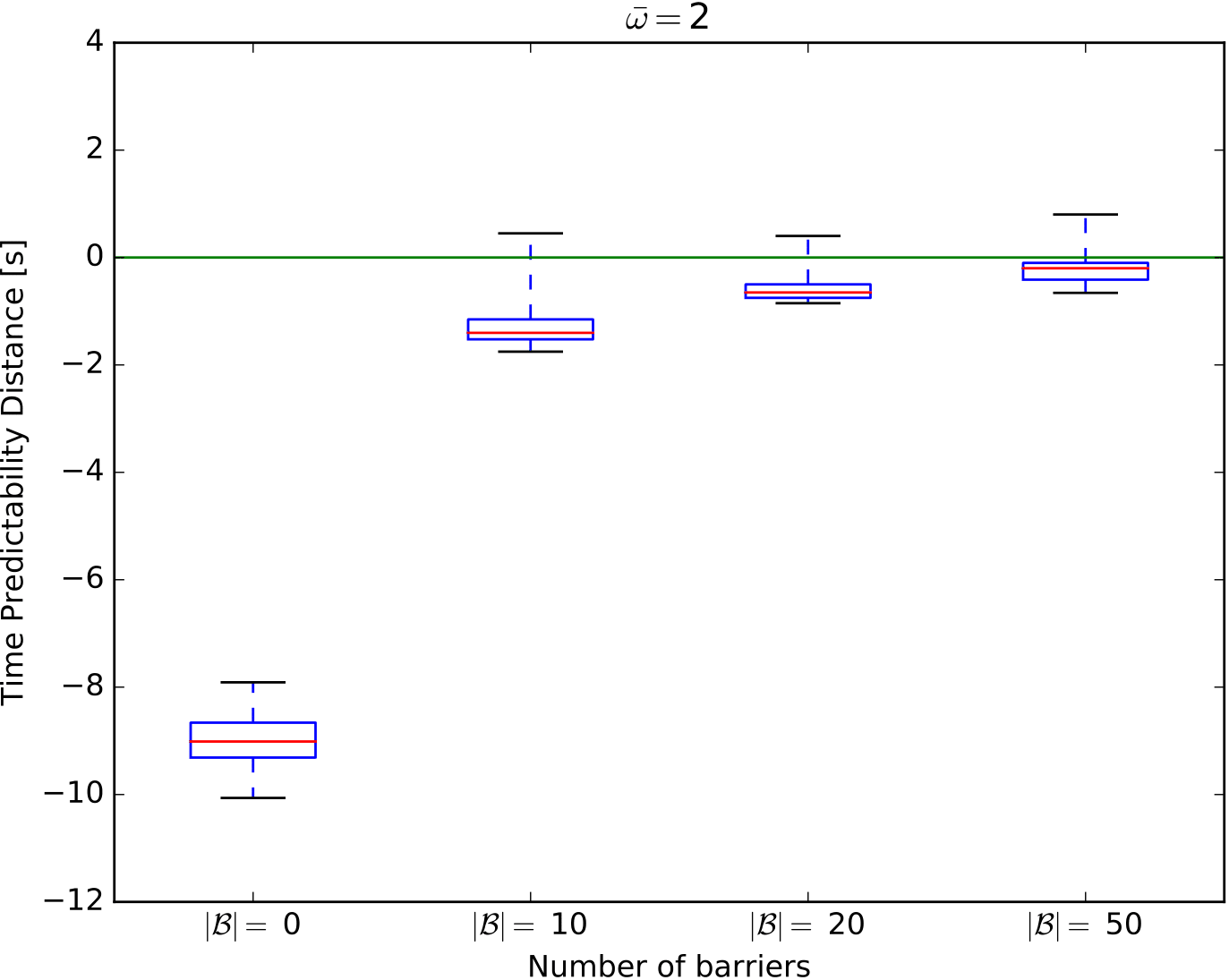}
\caption{Predictability distances with $\bar p = 0.6$ and  $\bar \omega = 2$.}
\end{subfigure}
\begin{subfigure}[t]{0.49\columnwidth}
\includegraphics[width=\columnwidth]{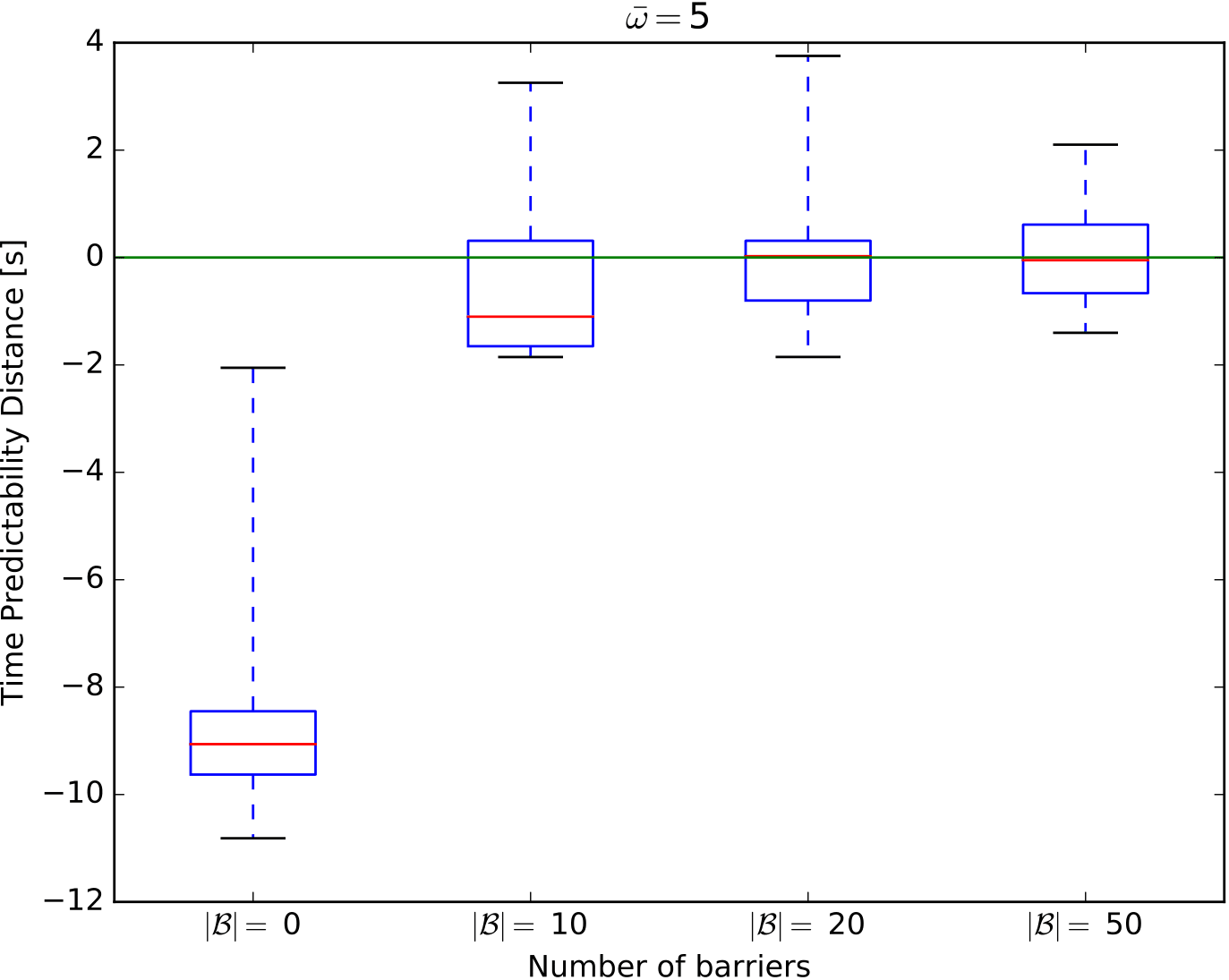}
\caption{Predictability distances with $\bar p = 0.6$ and  $\bar \omega = 5$.}
\end{subfigure}
\caption{Plotbox for predictability distance for Example~\ref{PM.ex.dummy.timeline}. $\mathcal{B} = 0$ refers to the unsynchronized parallel execution.}
\label{PM:fig:timeline}
\end{figure}

Note how, without synchronization, the predictability distance is negative. This is due to the fact that the action $\act_2$ has a progress that increases faster than the desired progress profile.
\subsubsection{Consideration on Safeguarding BTs}
\label{sec:PM:safeguarding}
For each BT of the Examples~\ref{PM.ex.dummy} and ~\ref{PM.ex.dummy.delta} above, the progress difference between $x_k$ and $x_{k+1}$ cannot exceed $\alpha + 2\bar \omega$. This characteristic is similar to the concept of  safeguarding BTs (see Definition~\ref{bg.def.safeguarding}). If the progress function is a monotonic increasing function, then the step length can hint on the synchronization performance. 
Consider the case in which $p(x_k)=x_k$, the BTs of the examples above are safeguarding with respect to the step length $d = \alpha + 2\bar \omega $.  Intuitively, given the results in this section, a safeguarding BT with smaller step length have better performance.

\newpage
\section{Experimental Evaluation}
\label{EE}

In this section, we present the experimental evaluation in four different realistic scenarios, one highlighting the absolute progress synchronization (introduced in Section~\ref{PM:AS}), one highlighting the relative progress synchronization (introduced in Section~\ref{PM:RS}), one showing how to use the proposed framework to impose coordination between perpetual actions, and one predictability (introduced in Section~\ref{PM:IP}). To collect statistically-significant data, for we ran each experiment 100 times and we show, plotboxes for the value of the measure described in Section~\ref{sec:performance}. The experimental results support the analysis done in Section~\ref{sa}.
A video showing the execution of the experiments is publicly available.\footnote{\url{https://youtu.be/eDrZp7n3y-s}}

%
%
%
%

\begin{experiment}[Absolute Progress Synchronization]
\label{EE:ex:Seek}
An object-seeking robot has to detect and recognize possible objects, to clear them, on the floor of a hallway. The robot's behavior is described as the absolute parallel node of two actions: \emph{swipe} that makes the head periodically move sideways while scanning possible objects, and \emph{navigate} that moves the mobile base through the hallway.  The progress of both actions
increases in linear proportion to the part of the hallway navigated or swept. Whenever the robot finds an unidentified object, the swipe action stops moving the robot's head until the object is identified. To correctly execute the task, the two actions are synchronized with an absolute parallel synchronized node. Figures~\ref{EE:Fig:Seek:UnSync0}, \ref{EE:Fig:Seek:UnSync1}, and~\ref{EE:Fig:Seek:UnSync2} show the steps executed by the robot without a synchronization, whereas Figures~\ref{EE:Fig:Seek:Sync0}, \ref{EE:Fig:Seek:Sync1}, and~\ref{EE:Fig:Seek:Sync2} show the steps executed by the robot with synchronization. Figure~\ref{EE:fig:Seek:real:abs} shows the performance of the synchronized and unsynchronized execution in different settings.
\end{experiment}

\begin{figure}[h!]
\begin{subfigure}[t]{0.49\columnwidth}
\includegraphics[width=\columnwidth]{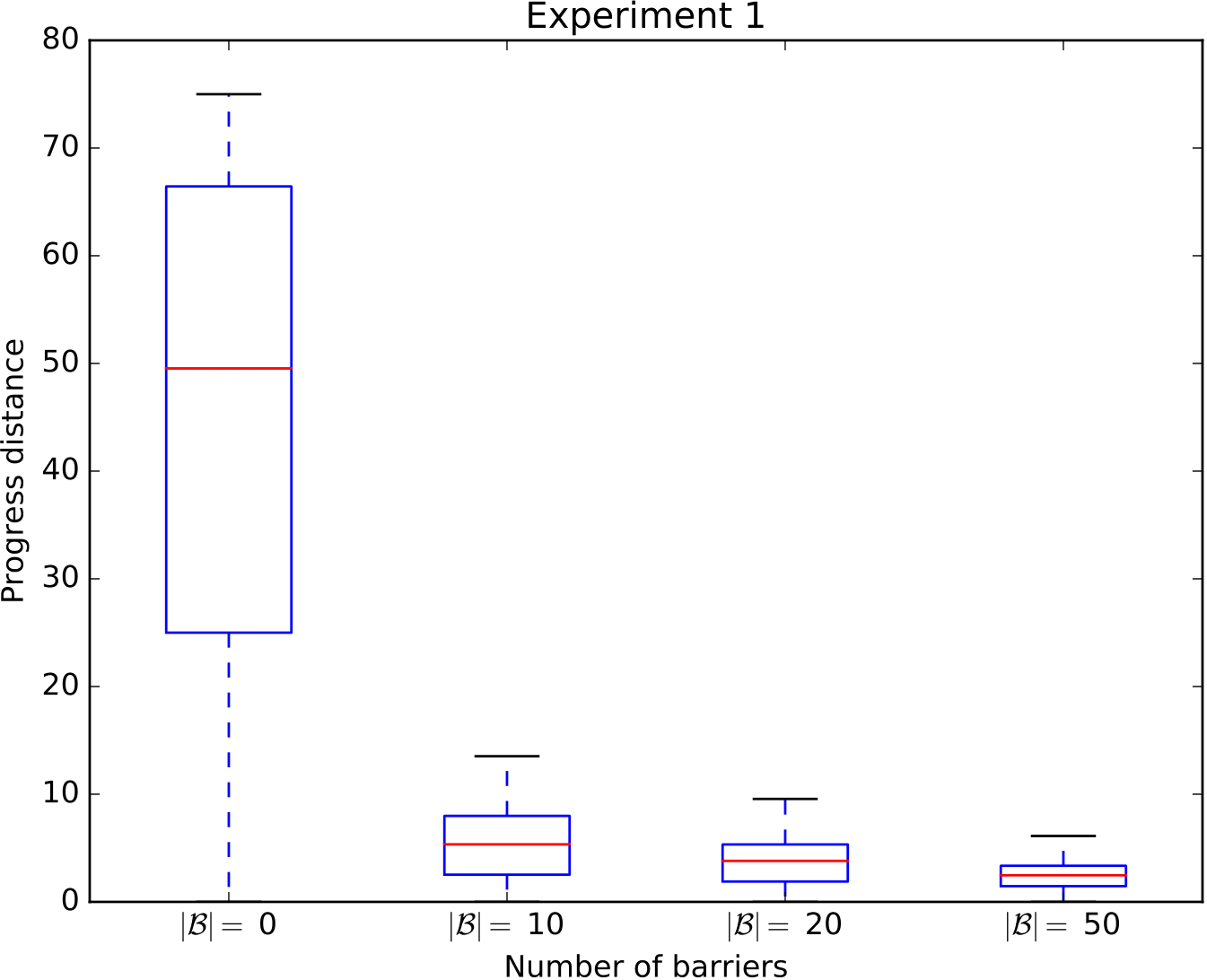}
\caption{Experiment~\ref{EE:ex:Seek}.}
\label{EE:fig:Seek:real:abs}
\end{subfigure}
\begin{subfigure}[t]{0.49\columnwidth}
\includegraphics[width=\columnwidth]{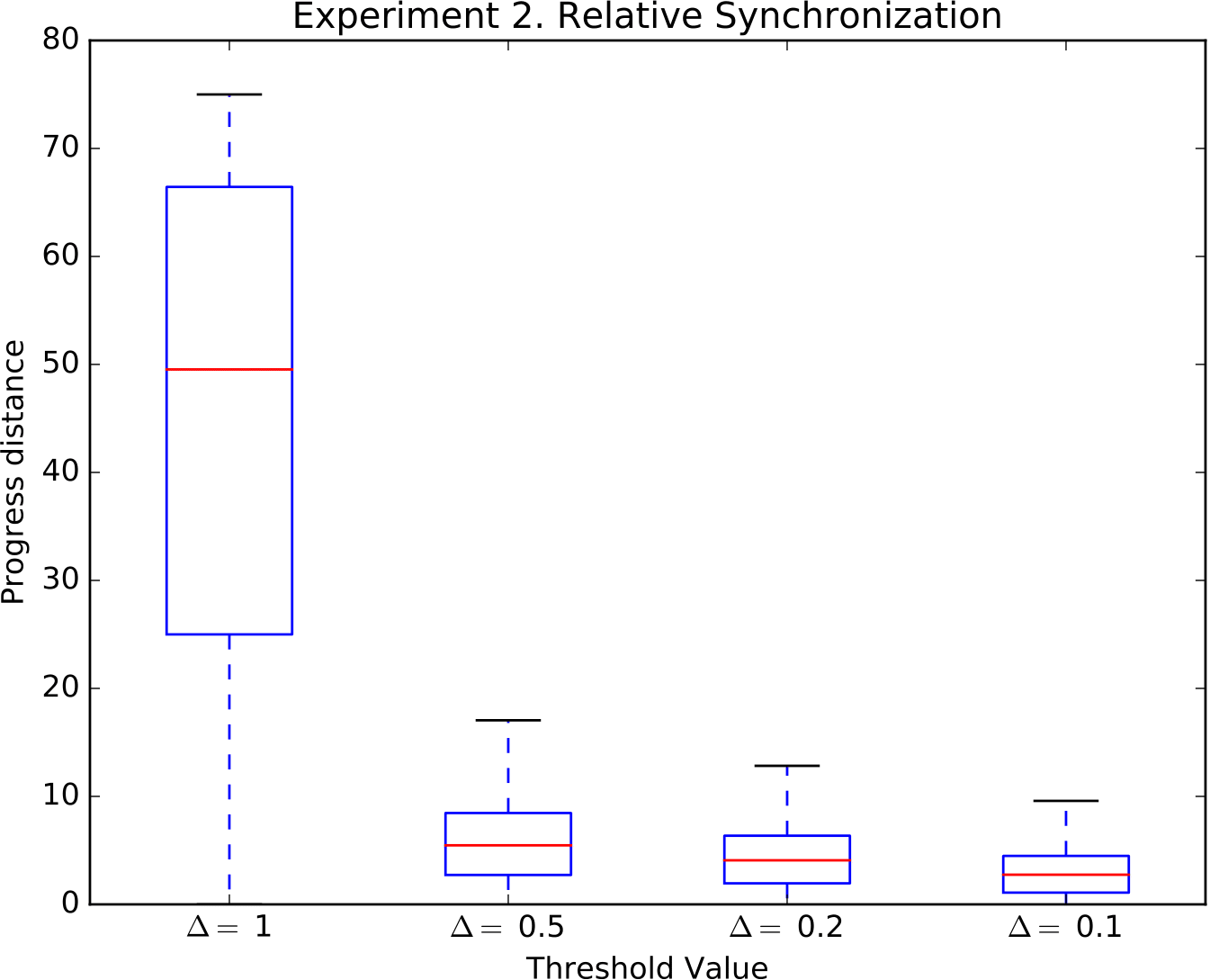}
\caption{Experiment~\ref{EE:ex:Seek.rel}.}
\label{EE:fig:Seek:real:rel}
\end{subfigure}
\caption{Plotbox of progress distance for Experiments~\ref{EE:ex:Seek} and~\ref{EE:ex:Seek.rel}. $\mathcal{B} = 0$ refers to the unsynchronized parallel execution.}
\label{EE:fig:Seek:real}
\end{figure}


\begin{experiment}[Relative Progress Synchronization]
\label{EE:ex:Seek.rel}
Consider the task on Experiment~\ref{EE:ex:Seek}. The actions \emph{swipe} and \emph{navigate} are now composed using relative synchronized parallel node. Figure~\ref{EE:fig:Seek:real:rel} shows the performance of the synchronized and unsynchronized execution with different settings.
\end{experiment}

\begin{remark}
At the design stage for single actions, we did not specify any speed for the head or mobile base movement. That gives freedom to the user to define any speed and reuse a pre-used action. However, the remark of the safeguarding BTs in Section~\ref{sec:PM:safeguarding} must be taken into account. 
\end{remark}

\begin{experiment}[Synchronization of Perpetual Actions]
\label{EE:ex:Cart}
A service robot is tasked to carry tools inside a workshop. When the tools are too many or too heavy, the robot uses a cart to carry them. The robot's behavior is described as a relative parallel node with two actions as children: \emph{hold cart} and \emph{follow path}.  The progress of both actions is $1$ whenever the error of, respectively, arm or base reference position is within a boundary, $0$ otherwise. 

While the robot is pushing the cart forwards, the cart may drift sideways, just like any ordinary cart.  To avoid rigidity, the arms' controllers are compliant, hence the drift of the cart would make the arms move with the cart. 
Whenever the cart drifts, the error of the action \emph{hold cart} increases. To avoid that the robot keeps moving while the cart drifts too much, the BT is a relative synchronized parallel node of the two actions.



Figures~\ref{EE:Fig:Cart:UnSync0}, \ref{EE:Fig:Cart:UnSync1}, \ref{EE:Fig:Cart:UnSync2}, and \ref{EE:Fig:Cart:UnSync3} show the steps executed by the robot without synchronization, whereas  Figures~\ref{EE:Fig:Cart:Sync0}, \ref{EE:Fig:Cart:Sync1}, \ref{EE:Fig:Cart:Sync2}, \ref{EE:Fig:Cart:Sync3} show the steps executed by the robot with synchronization. 
\end{experiment}

\begin{experiment}[Predictability]
\label{EE:ex:timeline}
An industrial robot has to perform some specific manipulation operations. Depending on the current tool used, some movements must follow a straight-line profile (same movement's speed throughout the execution) some other follow a sigmoid profile (the movements are first slow, then fast, then slow again). 
The robot's behavior is described by the BT in Figure~\ref{EE:fig:timeline:BT},  where the action \emph{do task} describes the manipulation task without any specific progress profile. 
Figures~\ref{EE:fig:timeline:straight} and \ref{EE:fig:timeline:sigmoid} shows the real and the desired progress profile in the two cases. 
\end{experiment}

\begin{figure}[h!]
\centering
\begin{subfigure}[t]{\columnwidth}
\centering
\includegraphics[width=\columnwidth]{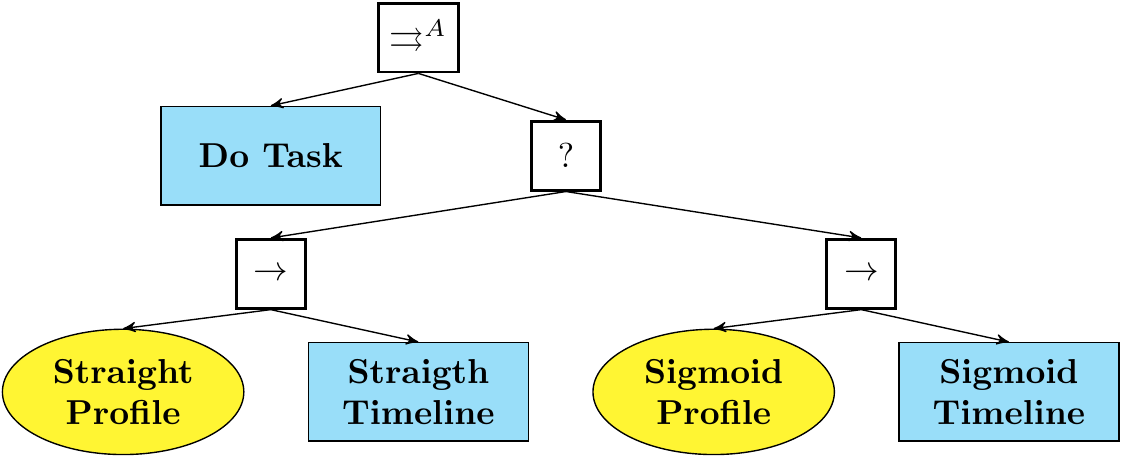}
\caption{BT for Experiment~3. The parallel node is an absolute synchronized parallel node with $50$ barriers.}
\label{EE:fig:timeline:BT}
\end{subfigure}
\vspace{1em}

\begin{subfigure}[t]{0.49\columnwidth}
\includegraphics[width=\columnwidth]{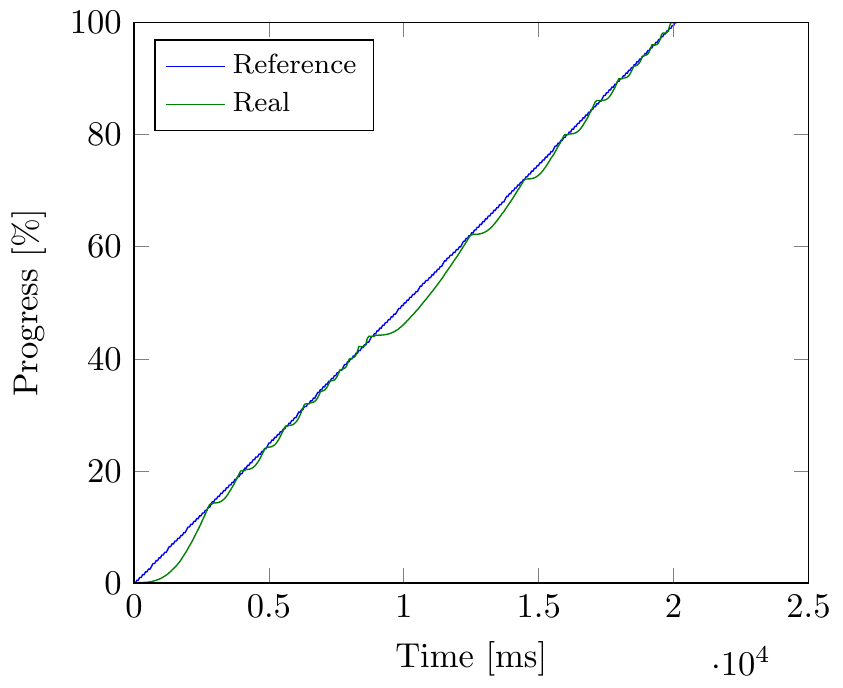}
\caption{Straight progress profile case.}
\label{EE:fig:timeline:straight}
\end{subfigure}
\begin{subfigure}[t]{0.49\columnwidth}
\includegraphics[width=\columnwidth]{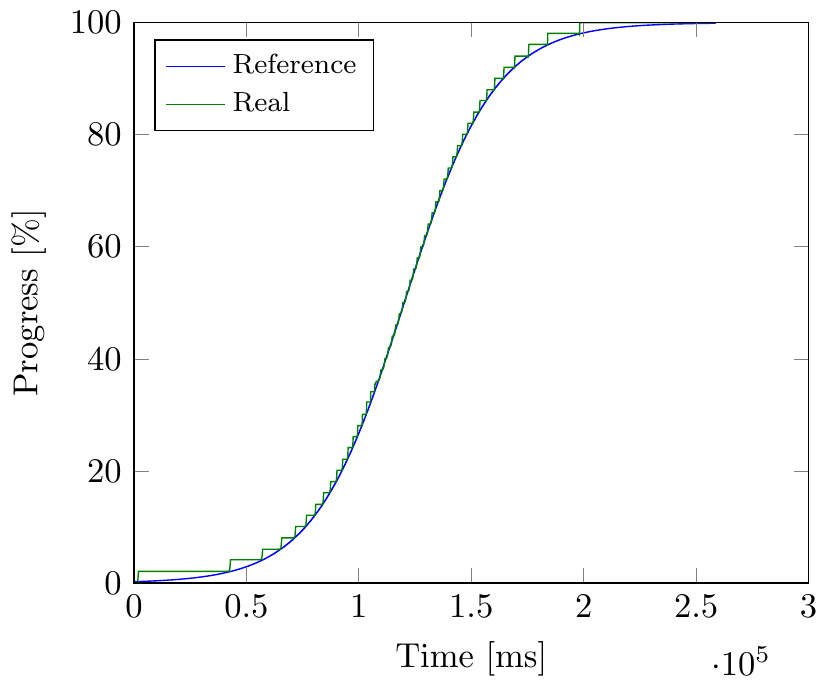}
\caption{Sigmoid progress profile case.}
\label{EE:fig:timeline:sigmoid}
\end{subfigure}
\caption{BT and progress profiles for Experiment~~\ref{EE:ex:timeline}.}
\label{EE:fig:timeline}
\end{figure}

\newpage
\section{Conclusions}
\label{sec:conclusions}

In this paper, we proposed two new BTs control flow nodes for progress synchronization with different synchronization policies, absolute and relative. We proposed measures to assess the synchronization between different sub-BTs and the predictability of robots execution. Moreover, we observed how design choices for the synchronization may affect the performance. Such observations are supported by experimental validation. 

We showed the applicability of our approach in a simulation system that allowed us to run the experiments several times in different settings to collect statistically-significant data.

%

\begin{figure}[b!]
\centering

\begin{subfigure}[t]{0.49\columnwidth}
\includegraphics[width=\columnwidth,trim={20cm 13cm 20cm 2cm},clip]{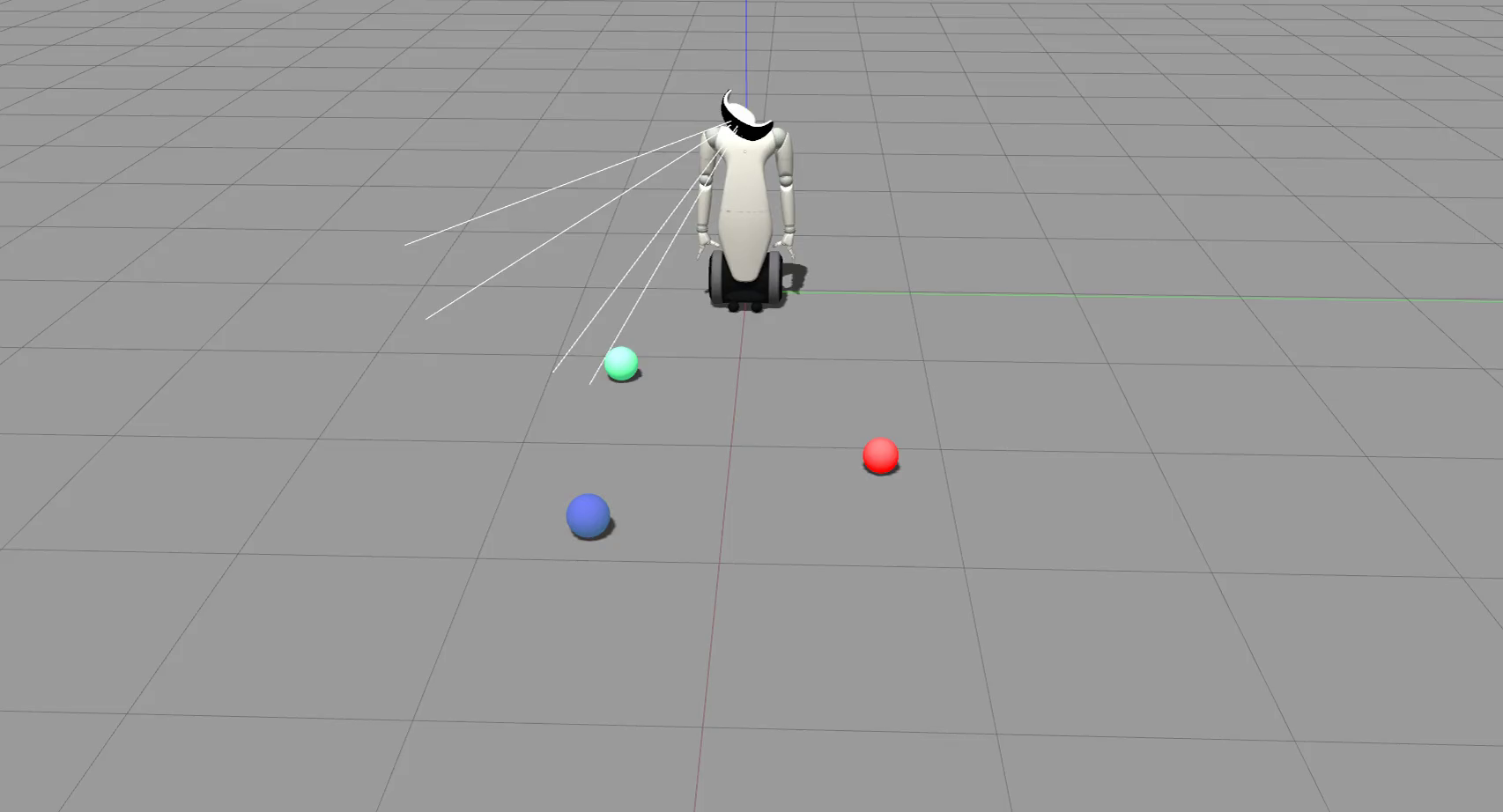}
\caption{[Unsync] The robot's head moves too fast, making the robot miss the first object.}
\label{EE:Fig:Seek:UnSync0}
\end{subfigure}
\vspace{1em}
\begin{subfigure}[t]{0.49\columnwidth}
\includegraphics[width=\columnwidth,trim={20cm 11cm 20cm 4cm},clip]{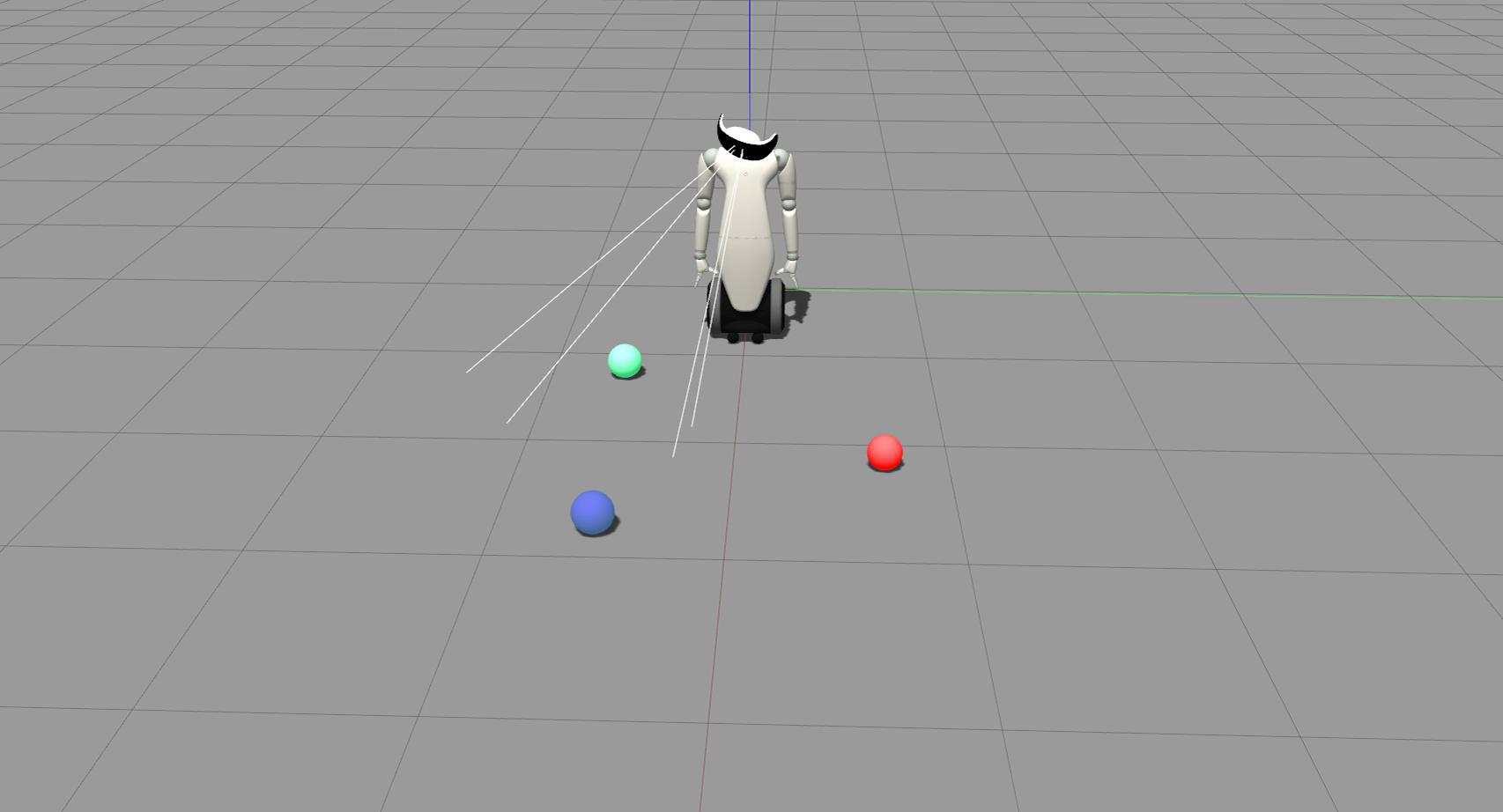}
\caption{[Sync] The robot finds an unidentified object on the floor. The head movement stops, so does the progress, the base movement is paused accordingly. \newline
}
\label{EE:Fig:Seek:Sync0}
\end{subfigure}

\vspace{1em}

\begin{subfigure}[t]{0.49\columnwidth}
\includegraphics[width=\columnwidth,trim={20cm 10cm 20cm 5cm},clip]{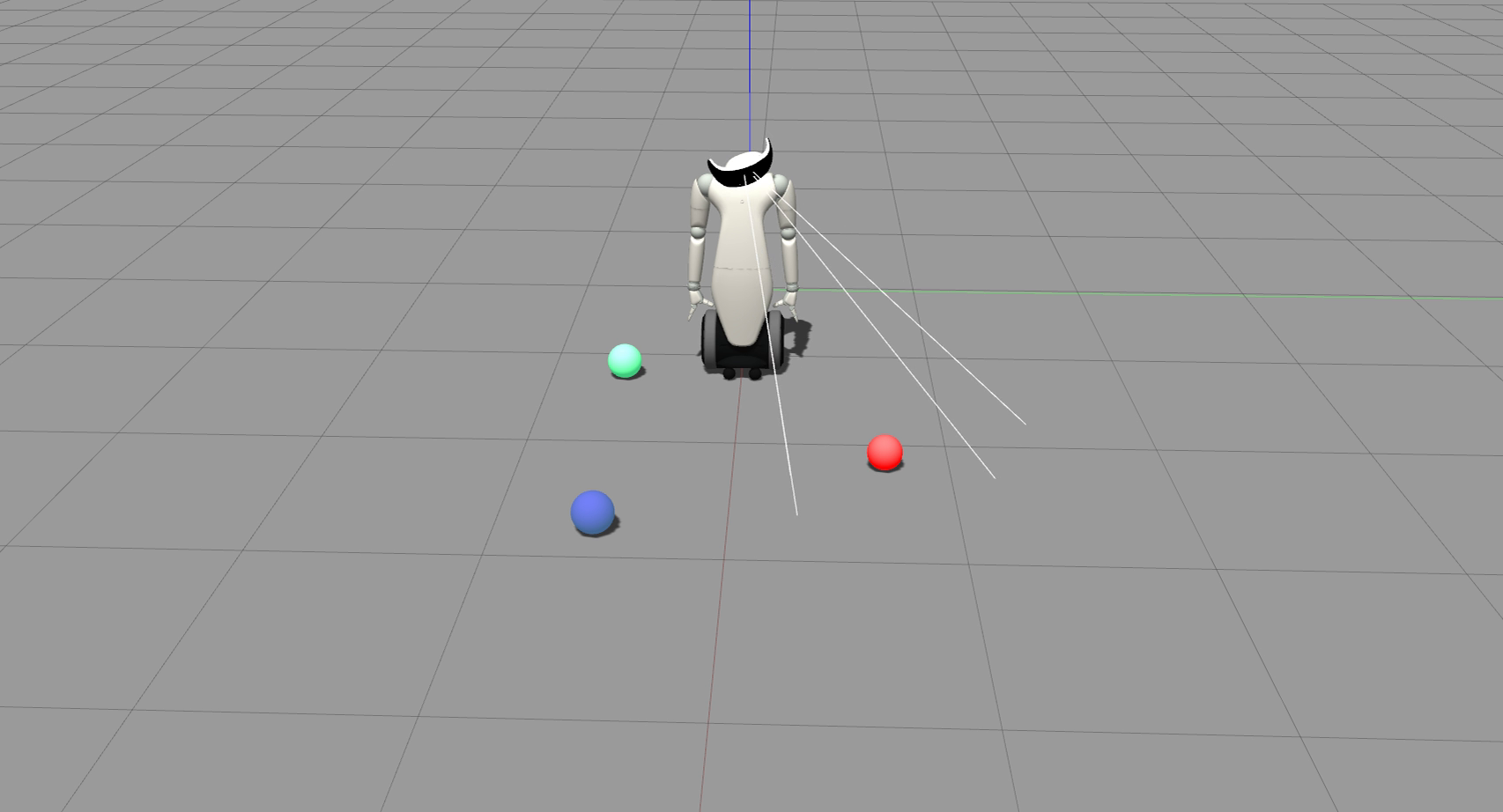}
\caption{[Unsync] The robot finds an unidentified object on the floor. The head movement stops, but the robot keeps moving, making the object be out of the robot's field of view.}
\label{EE:Fig:Seek:UnSync1}
\end{subfigure}
\begin{subfigure}[t]{0.49\columnwidth}
\includegraphics[width=\columnwidth,trim={20cm 10cm 20cm 5cm},clip]{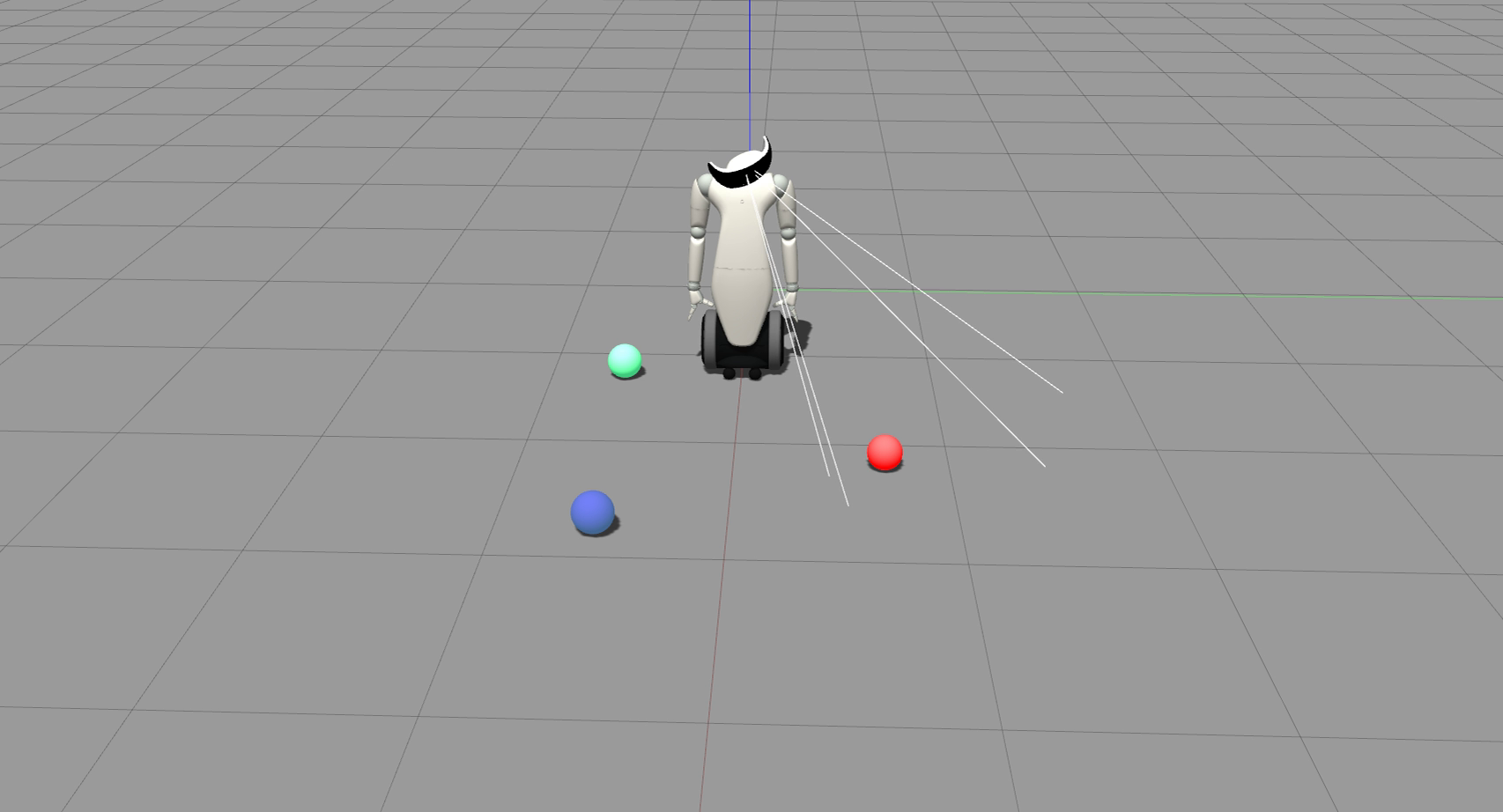}
\caption{[Sync] The robot finds another unidentified object on the floor. The head movement stops, so does the progress, the base movement is paused accordingly.}
\label{EE:Fig:Seek:Sync1}
\end{subfigure}
\vspace{1em}

\begin{subfigure}[t]{0.49\columnwidth}
 \includegraphics[width=\columnwidth,trim={20cm 10cm 20cm 5cm},clip]{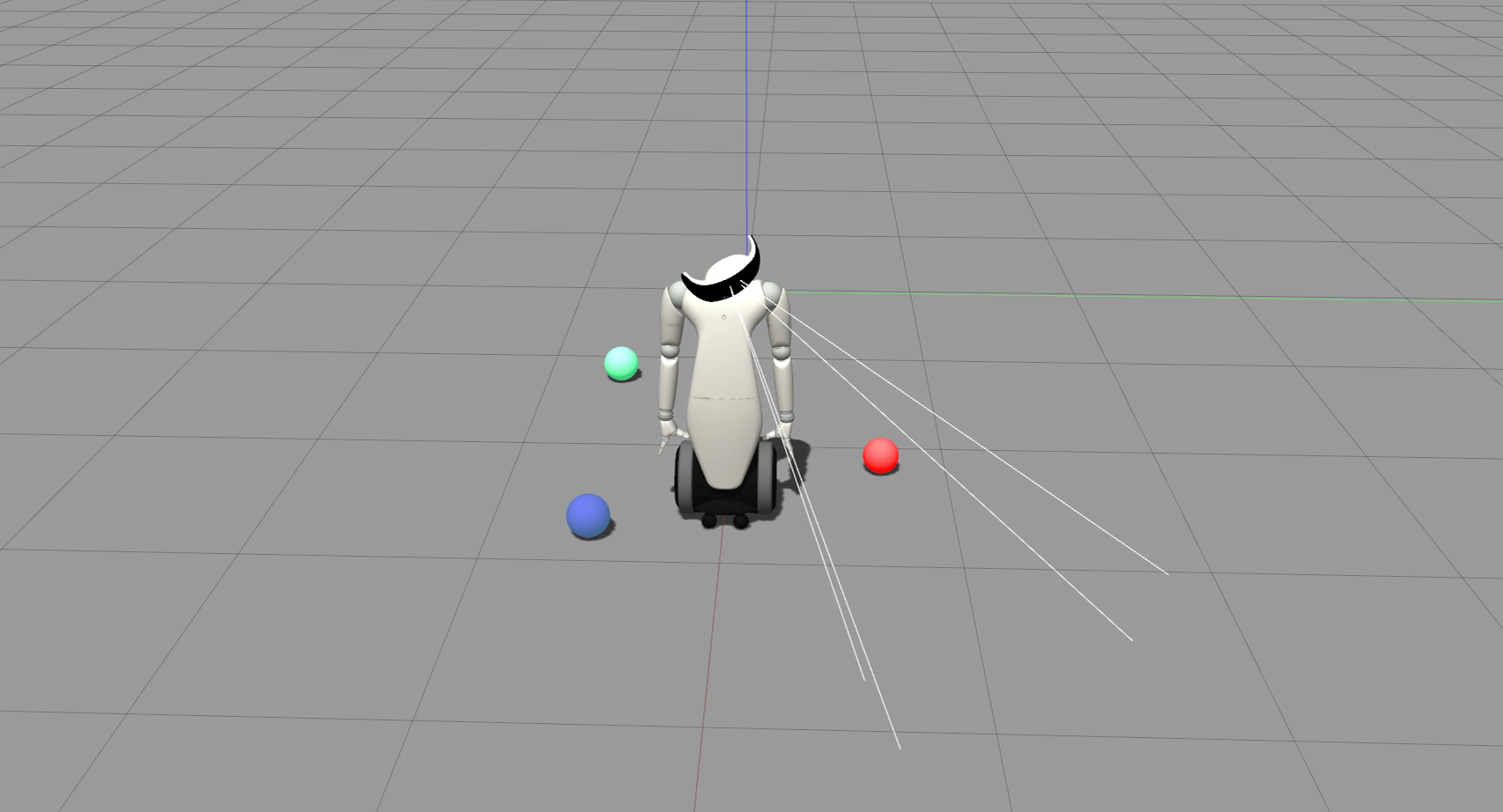}
\caption{[Unsync] The robot misses the other objects.}
\label{EE:Fig:Seek:UnSync2}
\end{subfigure}
\begin{subfigure}[t]{0.49\columnwidth}
\includegraphics[width=\columnwidth,trim={20cm 10cm 20cm 5cm},clip]{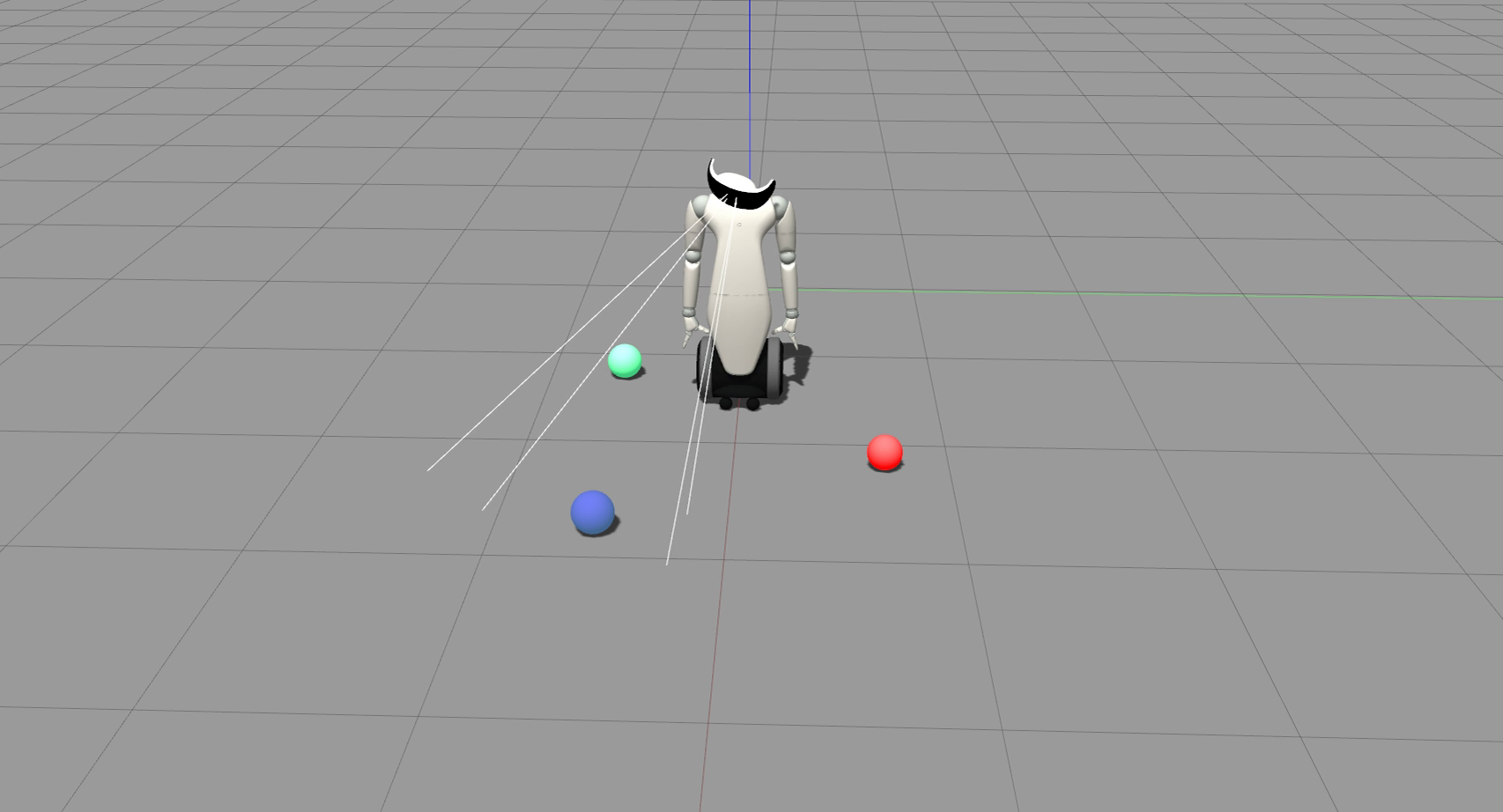}
\caption{[Sync] The robot recognized the object and resumes the head movement. The base movement is resumed as well.}
\label{EE:Fig:Seek:Sync2}
\end{subfigure}

\vspace{1em}

\caption{Execution steps related to Experiment~\ref{EE:ex:Seek}.}
\label{EE:fig:Seek}
\end{figure}

\begin{figure}[h!]
\centering

\begin{subfigure}[t]{0.49\columnwidth}
\includegraphics[width=\columnwidth,trim={15cm 5cm 25cm 10cm},clip]{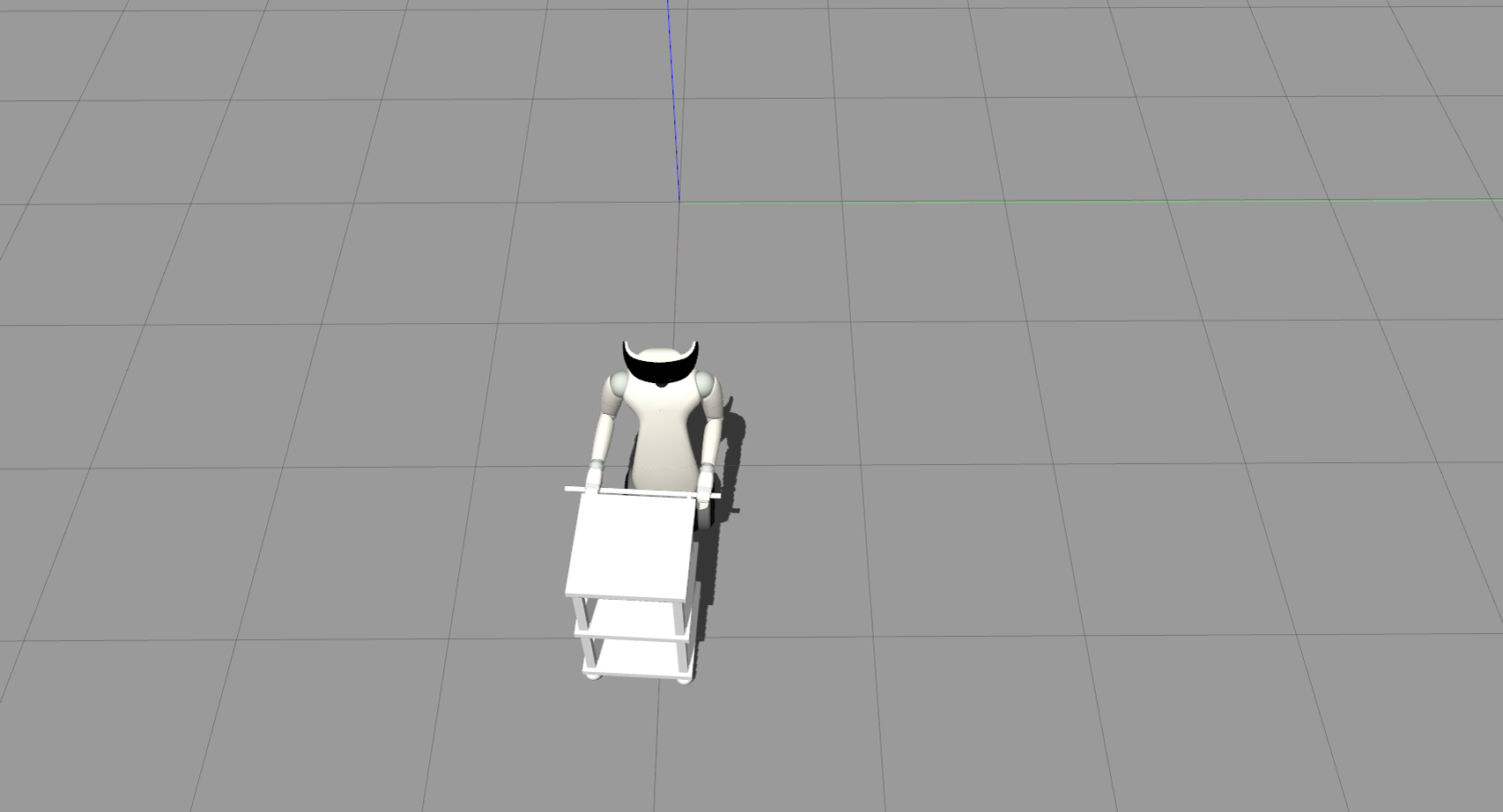}
\caption{[Unsync] The robot is holding the cart while moving. The cart drifts in such a way that the reference error crosses the threshold value. The parallel node keep sending ticks to the follow path action.}
\label{EE:Fig:Cart:UnSync0}
\end{subfigure}
\begin{subfigure}[t]{0.49\columnwidth}
\includegraphics[width=\columnwidth,trim={15cm 5cm 25cm 10cm},clip]{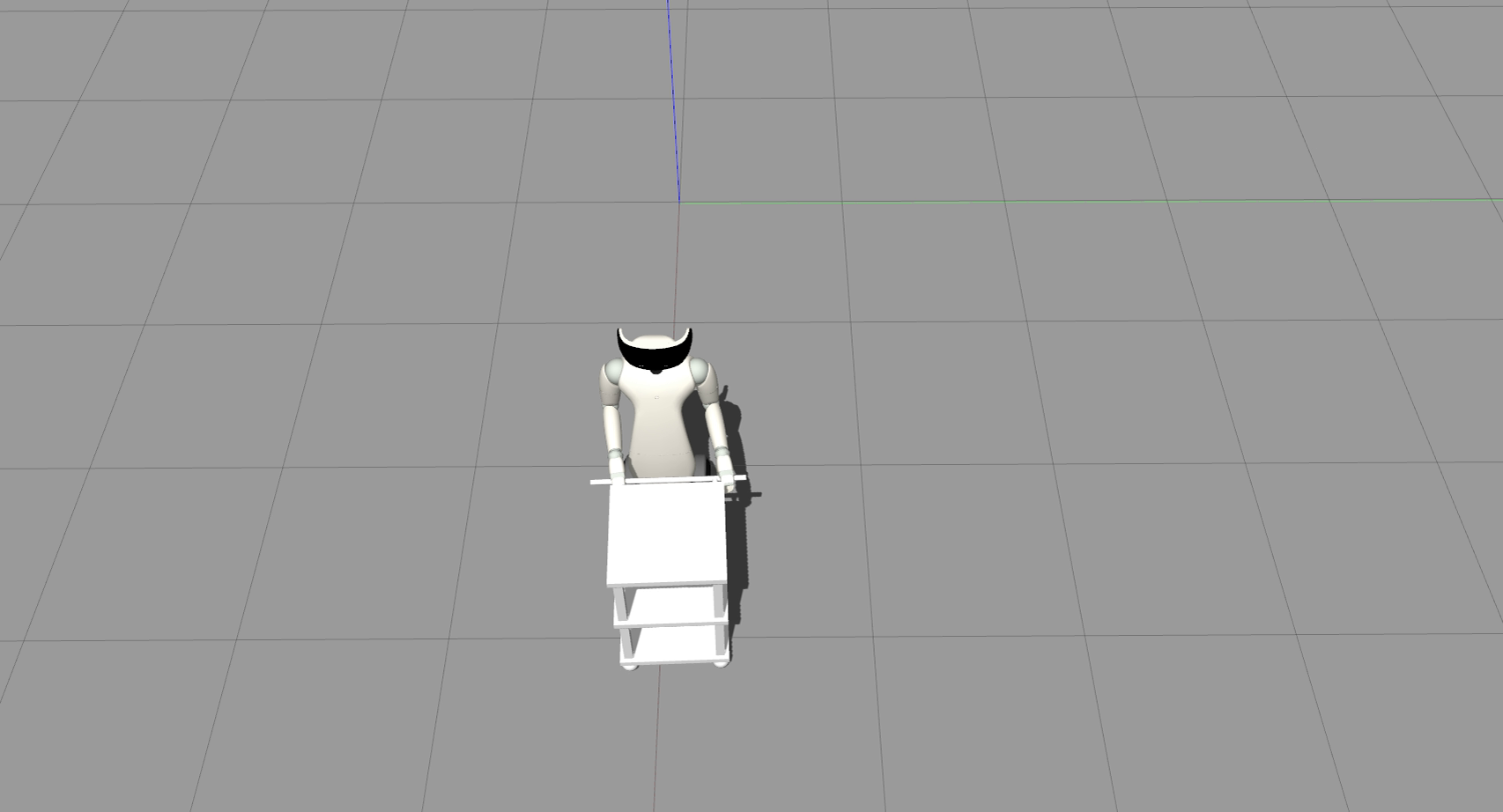}
\caption{[Sync] The robot is holding the cart while moving. The cart drifts in such a way that the reference error crosses the threshold value. The parallel node stops sending ticks to the follow path action and robot stops moving.}
\label{EE:Fig:Cart:Sync0}
\end{subfigure}
\vspace{1em}

\begin{subfigure}[t]{0.49\columnwidth}
\includegraphics[width=\columnwidth,trim={15cm 2cm 25cm 13cm},clip]{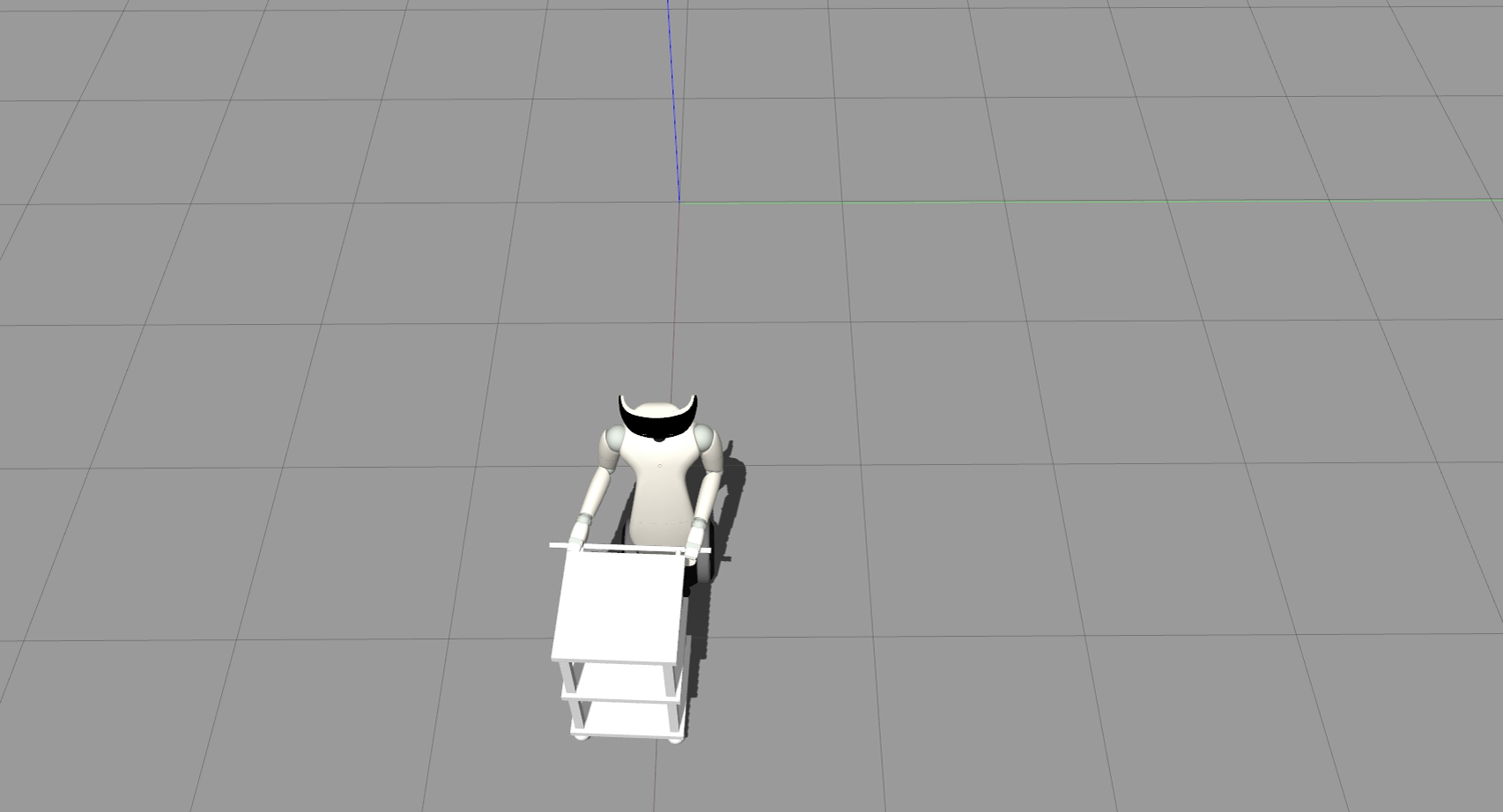}
\caption{[Unsync] The robot  does not stop moving its mobile base and the cart keeps drifting.}
\label{EE:Fig:Cart:UnSync1}
\end{subfigure}
\begin{subfigure}[t]{0.49\columnwidth}
\includegraphics[width=\columnwidth,trim={15cm 0cm 25cm 15cm},clip]{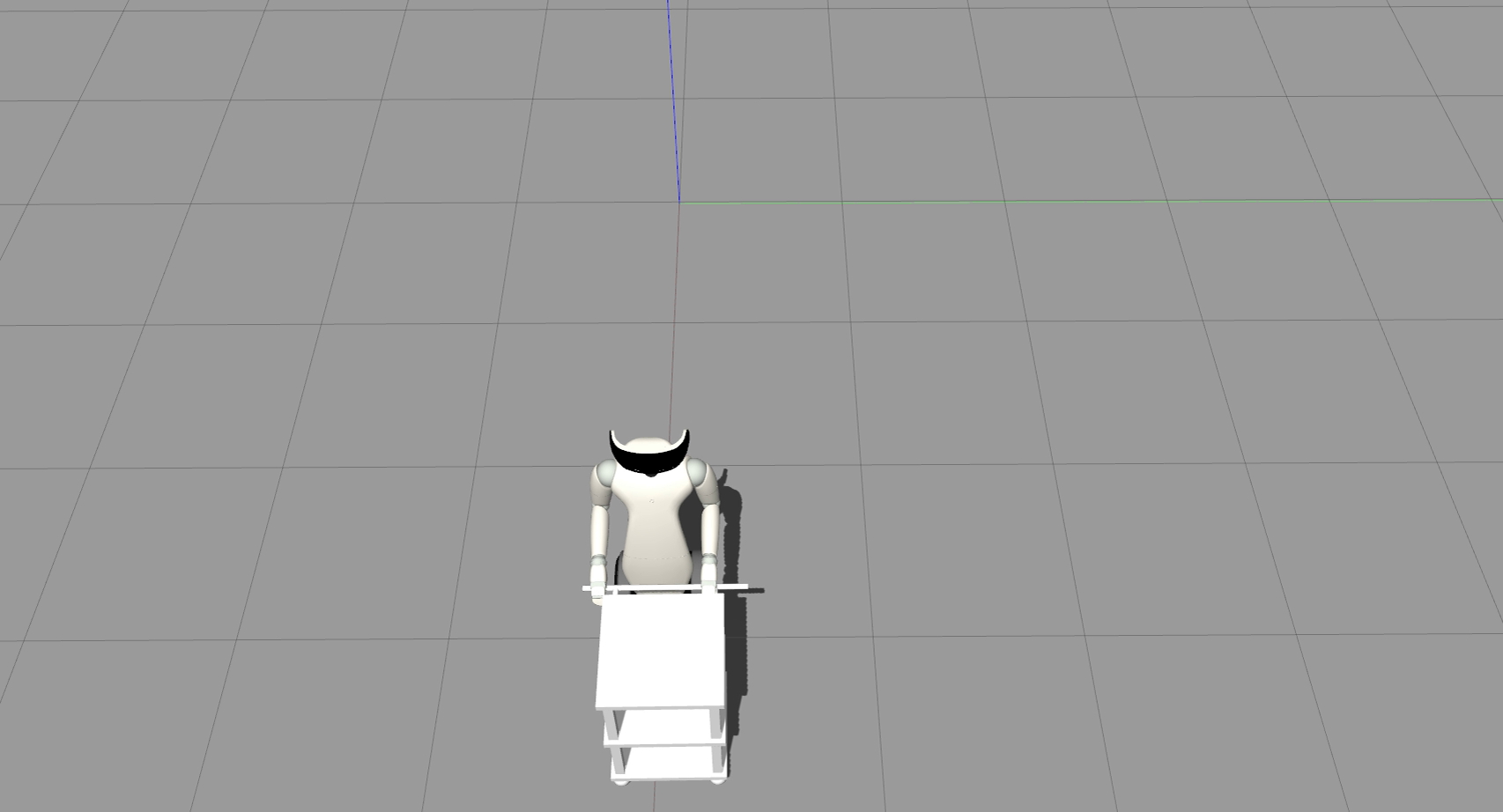}
\caption{[Sync] The robot aligns correctly the cart. The navigation resumes.\newline \newline
}
\label{EE:Fig:Cart:Sync1}
\end{subfigure}

\vspace{1em}
\begin{subfigure}[t]{0.49\columnwidth}
\includegraphics[width=\columnwidth,trim={15cm 8cm 25cm 7cm},clip]{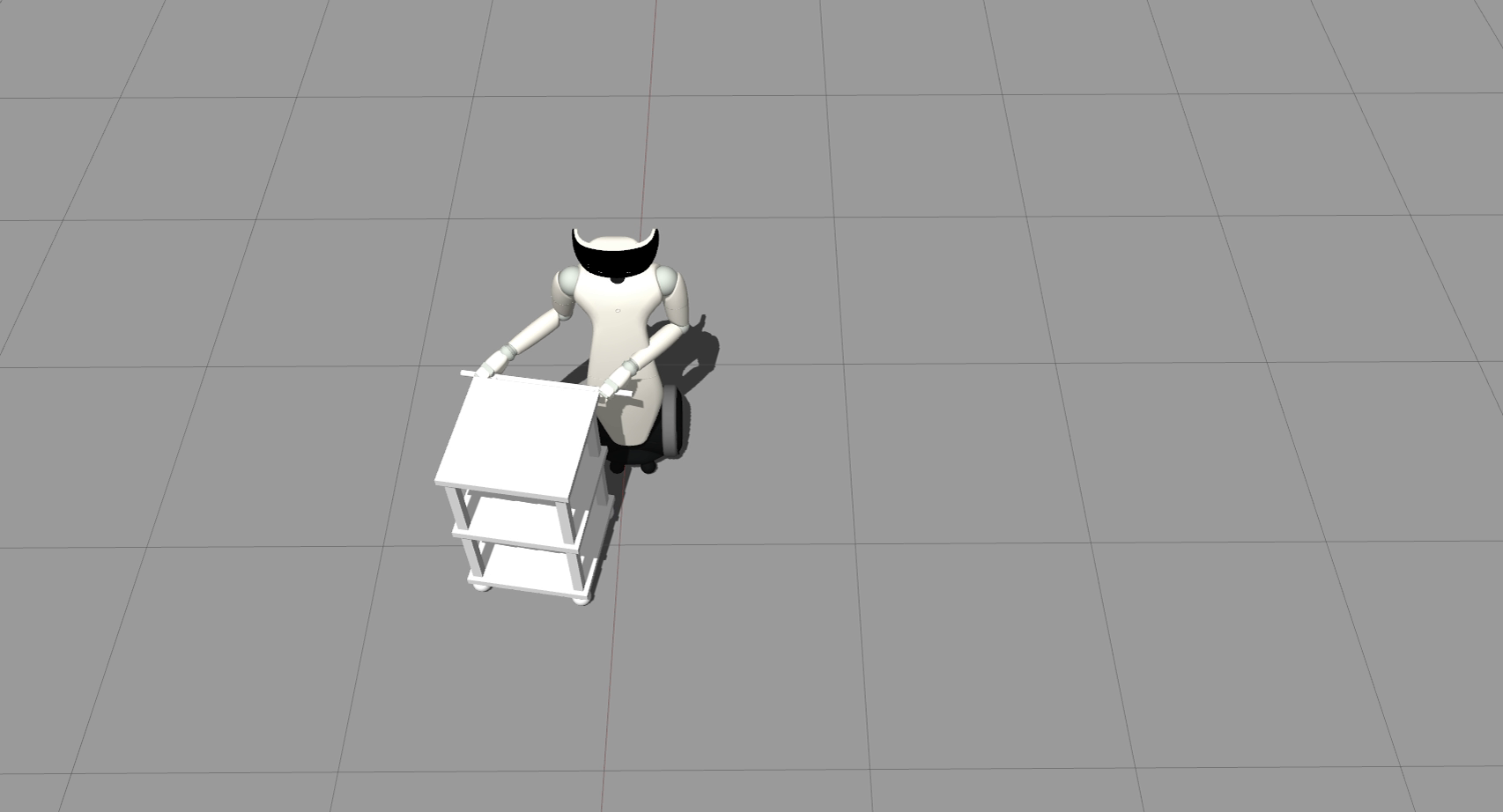}
\caption{[Unsync] The cart slips out from the robot's hand.}
\label{EE:Fig:Cart:UnSync2}
\end{subfigure}
\begin{subfigure}[t]{0.49\columnwidth}
\includegraphics[width=\columnwidth,trim={15cm 2cm 25cm 13cm},clip]{new_cart_unsync}
\caption{[Sync] The robot  does not stop moving its mobile base and the cart keeps drifting.}
\label{EE:Fig:Cart:Sync2}
\end{subfigure}
\vspace{1em}

\begin{subfigure}[t]{0.49\columnwidth}
\includegraphics[width=\columnwidth,trim={15cm 8cm 25cm 7cm},clip]{R1_cart_unsync_1}
\caption{[Unsync] The cart slips out from the robot's hand.}
\label{EE:Fig:Cart:UnSync3}
\end{subfigure}
\begin{subfigure}[t]{0.49\columnwidth}
\includegraphics[width=\columnwidth,trim={15cm 0cm 25cm 15cm},clip]{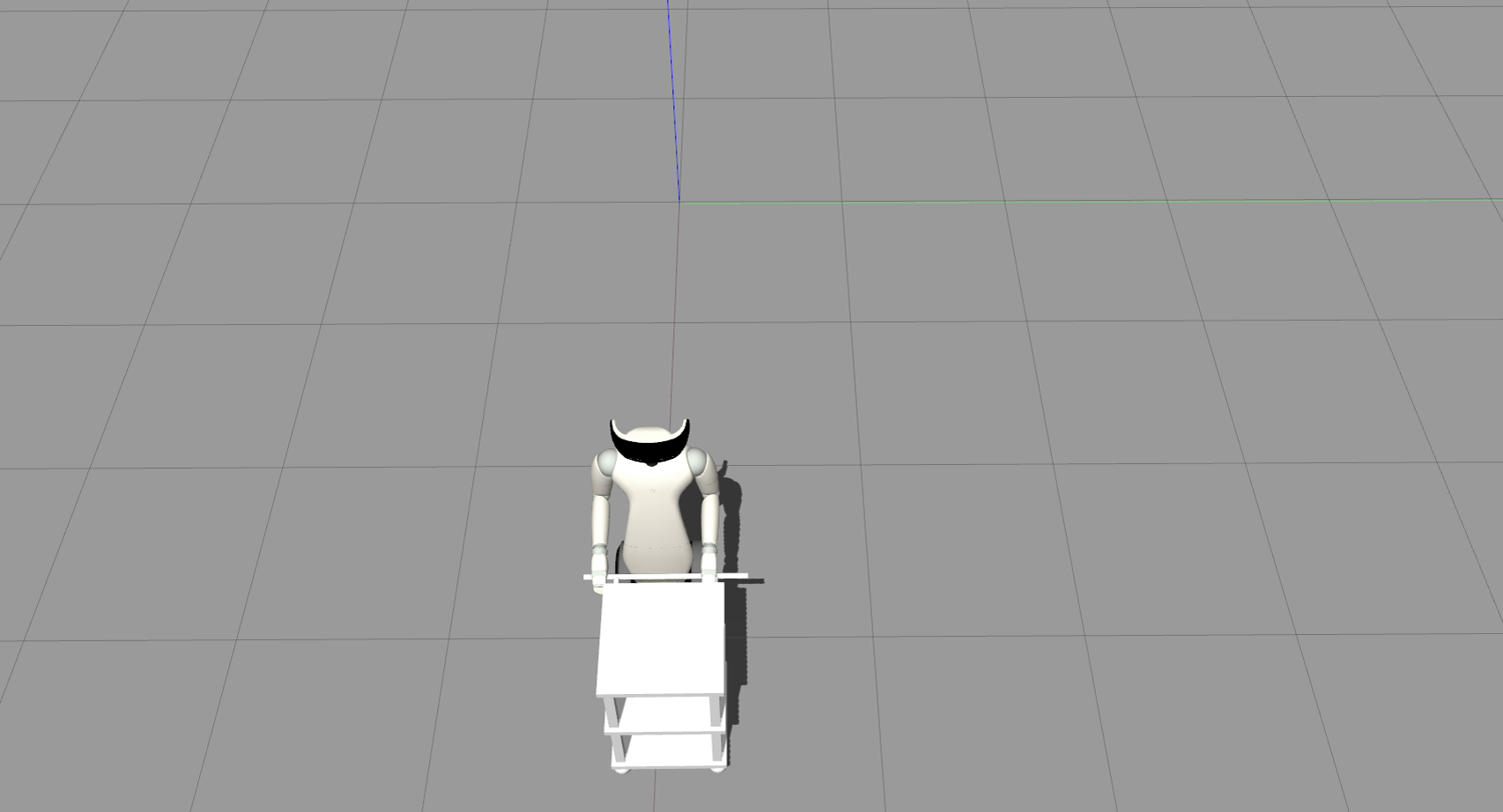}
\caption{[Sync] The robot keeps pushing correctly the cart.
\label{EE:Fig:Cart:Sync3}
\newline
\newline
}
\end{subfigure}

\vspace{1em}

\caption{Execution steps related to Experiment~\ref{EE:ex:Cart}.}
\label{EE:fig:Cart}
\end{figure}


\newpage
\vspace*{1em}

\bibliographystyle{IEEEtran}
\bibliography{behaviorTreeRefs}
\end{document}